\documentclass[conference,compsoc]{IEEEtran}
\IEEEoverridecommandlockouts  

\usepackage{hyperref}
\usepackage{times}
\usepackage{soul}
\usepackage{url}
\usepackage{caption}

\usepackage{graphicx}
\usepackage{amsmath}
\usepackage{amsthm}
\usepackage{booktabs}
\usepackage{algorithm}
\usepackage{algorithmic}
\usepackage{enumitem}
\usepackage{lipsum}

\DeclareMathOperator*{\argmin}{arg\,min}
\usepackage{amsfonts}
\usepackage{epstopdf,epsfig}
\usepackage{xcolor}
\usepackage{balance}
\usepackage{colortbl}
\usepackage{pbox}
\usepackage{tabularx}

\usepackage{subcaption}
\usepackage{tabularx,booktabs}
\usepackage{savesym}
\savesymbol{checkmark}
\usepackage{dingbat}
\usepackage{diagbox}
\usepackage{mathtools} 
\usepackage{booktabs} 
\usepackage{tikz} 
\newtheorem{theorem}{Theorem}

\newtheorem{lemma}{Lemma}
\newcommand{\changed}[1]{{\color{black}{#1}}}

\begin{document}

\title{\LARGE \bf Towards a Game-theoretic Understanding of Explanation-based Membership Inference Attacks}

\newcommand{\tudaAffiliation}{}

\author{
\IEEEauthorblockN{Kavita Kumari${^\dagger}{^*}$, Murtuza Jadliwala${^\ddagger}$, Sumit Kumar Jha${^\dagger}{^\dagger}$, Anindya Maiti${^*}{^*}$}
\IEEEauthorblockA{\textit{$^\dagger$Technical University of Darmstadt, $^\ddagger$The University of Texas at San Antonio}, \\
$^{\dagger\dagger}$Florida International University, $^{**}$University of Oklahoma\\
}
}
\maketitle

\def\thefootnote{*}\footnotetext{Work done while the author was affiliated with The University of Texas at San Antonio.}
%-------------------------------------------------------------------------------
\renewcommand*{\thefootnote}{\arabic{footnote}}

%\iffalse
\begin{abstract}
    %\vspace{-0.2cm}
Model explanations improve the transparency of black-box machine learning (ML) models and their decisions; however, they can also be exploited to carry out privacy threats such as membership inference attacks (MIA). Existing works have only analyzed MIA in a single ``what if" interaction scenario between an adversary and the target ML model; thus, it does not discern the factors impacting the capabilities of an adversary in launching MIA in repeated interaction settings. \changed{Additionally, these works rely on assumptions about the adversary’s knowledge of the target model’s structure and, thus, do not guarantee the optimality of the predefined threshold required to distinguish the members from non-members.}
%\Hossein{What are the drawbacks?}
%This paper studies explanation-based threshold attacks, where an adversary attempts to launch MIA using explanations variance by repeatedly interacting with the system (target ML model and its associated explanation method). 
In this paper, we delve into the domain of explanation-based threshold attacks, where the adversary endeavors to carry out MIA attacks by leveraging the variance of explanations through iterative interactions with the system comprising of the target ML model and its corresponding explanation method.
%The adversary's goal is to utilize historical (explanation) information to determine an appropriate explanation variance threshold that can be employed to determine the (training) membership of a set of data points. 
We model such interactions by employing a continuous-time stochastic signaling game framework. \changed{In our framework, an adversary plays a stopping game, interacting with the system (having imperfect information about the type of an adversary, i.e., honest or malicious) to obtain explanation variance information and computing an optimal threshold to determine the membership of a datapoint accurately. First, we propose a sound mathematical formulation to prove that such an optimal threshold exists, which can be used to launch MIA.}
%In this game, on the one hand, 
%There is a strategic interplay here between the adversary and the model that needs to be formally modeled in order to understand the feasibility conditions for such attacks.
%the adversary must strategically decide when to stop querying the target model in order to reach the target variance,
%\mj{can we replace ``assigned target" with something else? ``target variance"?}
%all without being detected. On the other hand, the system which has imperfect information about the type of end-user (malicious or honest) it is interacting with, 
%(i.e., it could be interacting with an honest user or a malicious adversary) 
%must strategically decide based on its Bayesian belief (about the end-user's type), how much information (in the form of noisy explanations) it should share, and when to stop sharing to prevent such MIAs. 
%This paper models this strategic interaction by means of a continuous-time stochastic Signaling game framework, and 
Then, we characterize the conditions under which a unique Markov perfect equilibrium (or steady state) exists in this dynamic system. 
%\Hossein{in this setup? - confusing!}.
By means of a comprehensive set of simulations of the proposed game model, we assess different factors that can impact the capability of an adversary to launch MIA in such repeated interaction settings.

\end{abstract}

\section{Introduction}
\label{sec:Intro}
 %\vspace{-0.2cm}

\changed{Membership Inference Attacks (MIAs) have been extensively studied in the literature, where adversaries analyze machine learning (ML) models to formulate attacks aimed at discerning the membership of specific data points. One prevalent approach involves classifying a training data record with high confidence while classifying a test data record with relatively lower confidence. This behavior of ML models allows attackers to distinguish members from non-members of the training dataset. MIAs can be classified into binary classifier-based attack approaches \cite{shokri2017membership, salem2018ml, yeom2018privacy, hui2021practical, nasr2019comprehensive, melis2019exploiting, leino2020stolen}, metric-based attack approaches \cite{long2017towards, salem2018ml, yeom2018privacy, song2021systematic}, and differential comparisons-based attacks \cite{hui2021practical}.

ML models often make 
%have been deployed in various industries, where these models make 
critical data-based decisions in various applications \cite{bhavsar2017machine, ahmad2018interpretable}, however due to the complex and black-box nature of these models, understanding the underlying reasons behind model decisions is often challenging. This has led to the development of a variety of model explanation techniques and toolkits \cite{ribeiro2016should, lundberg2017unified, shrikumar2017learning, arras2017relevant, jacovi2018understanding, ancona2018towards, GoogleMLInt, microsoftMLInt}. Simultaneously, explanations also expose an attack surface that can be exploited to infer private model information \cite{shokri2021privacy} or launch adversarial attacks against the model \cite{ignatiev2019relating, slack2020fooling}.
This work focuses specifically on metric-based MIAs, particularly explanation-based MIAs \cite{shokri2021privacy}. Existing attacks have suggested that points with maximum prediction confidence may act as non-member points \cite{salem2018ml} or have utilized threshold values through the shadow-training technique \cite{shokri2017membership, song2021systematic, yeom2018privacy}. However, one limitation of these approaches is the uncertainty surrounding the optimality of the computed threshold, and our main motivation in this work is to determine the optimal threshold for successfully launching a MIA. 
%with 100\% accuracy
Specifically, this work aims to theoretically guarantee the existence of an optimal threshold that adversaries can compute to launch MIA by exploiting state-of-the-art Explainable Artificial Intelligence (XAI) techniques for model explanations.

Adversaries in explanation-based MIA attacks leverage model explanations to infer the membership of target data points. Shokri et al. \cite{shokri2021privacy} explored the feasibility of explanation-based MIA attacks, demonstrating that backpropagation-based explanation variance, specifically the gradient-based explanation vector variance, can confirm the membership of target data samples when compared to a pre-defined threshold. In this work, we further analyze how the variance of explanations changes in the repeated query framework from an adversary to the system.  \changed{Here, the repeated play is the iterative interaction between the system (comprising ML model and explanation method) and an adversary, where an adversary sends repeated queries (formulated utilizing the historical information) to the system to inch closer to its goal of finding the optimal threshold of the explanation variance.}

In the context of metric-based MIAs, decisions regarding membership inference are made by calculating metrics on prediction vectors and comparing them to a preset threshold. Thus, it is a critical challenge in these threshold attacks to determine how to obtain or compute such a variance threshold. This problem can also be extrapolated to explanation-based MIA threshold attacks \cite{shokri2021privacy}. While it is straightforward to compute an optimal threshold if the training set membership is known \cite{long2017towards, salem2018ml, yeom2018privacy, song2021systematic}, the question arises: how can an explanation-based threshold attack be executed when an adversary lacks knowledge of the model and its training process? This poses a significant challenge that needs to be addressed to enhance the security of ML models against membership inference attacks.}

One strategy for an adversary to achieve this objective involves iteratively interacting with the target system to compute the explanation variance threshold. This adversarial play introduces multiple crucial questions: \emph{what is the optimal duration for which the adversary should interact with the target system? Can the target system identify such malicious interactions in time to prevent MIA? How can the target system adeptly and strategically serve both honest and malicious users in this context?} \changed{In this context, while the honest user can formulate queries similar to the malicious user, the emphasis lies on the malicious user's intention to initiate MIA. Thus, the malicious user is actively attempting to launch MIA. While the honest user, who may inadvertently employ similar strategies, is trying to obtain the reasoning of the model prediction for the queries it sent to the system. Thus, the value of an explanation for an honest end-user is based on its relevance, explaining the model's decision for the query. This relevance, inherently linked to the explanation's variance, varies accordingly. However, a malicious end-user evaluates an explanation's value based on the information it contains for potential exploitation in launching MIAs.}

Intuitively, the duration, pattern, and structure of such repeated interactions could impact the degree of private information disclosure by the system. Nevertheless, the current comprehension of this phenomenon is insufficient, particularly when considering the presence of a strategic adversary whose goal is to minimize the attack cost and path towards undermining the system's privacy and a strategic system who is aiming to prevent this without having full knowledge of the nature of the end-user (adversarial or non-adversarial) engaged in the interaction.

In this paper, we aim to bridge this research gap by utilizing a formal approach to model the strategic interactions between an adversary and an ML system via game theory. Specifically, we employ a \emph{continuous-time stochastic signaling game} framework to capture the complexities of the interaction dynamics. \changed{We have opted for a stochastic game to model repeated interaction between two agents, the adversary and the ML model, where an adversary makes an optimal control decision in each interaction instance, i.e., to continue or stop the process of sending queries and (their) explanations, respectively. Such problems involving optimal control are usually modeled with \emph{Bellman's} equation and solved using optimization techniques such as \emph{Dynamic Programming}. 
%Thus, the motivation behind utilizing Geometric Brownian Motion (GBM) to model explanation variance is that GBM excels in scenarios where agents may make decisions based on continuous and evolving information. 
We model explanation variance using a Geometric Brownian Motion (GBM) stochastic process because GBM excels in modeling scenarios where agents make decisions based on continuous and evolving information. Additionally, GBM's capacity to integrate historical data makes it more suitable for capturing strategic interactions in situations where past actions significantly influence current decisions. Lastly, as mentioned before, the value of the information of an explanation is contained in the variance of the explanation; thus, modeling in discrete time would have lost this vital information. Since the GBM framework is advantageous to model continuous and evolving information, modeling in continuous time (and using GBM) helps the system and the end-user to retain more information. Therefore, it ensures the soundness of the mathematical proof of an optimal threshold.}

To the best of our knowledge, no previous research has explored the use of repeated gradient-based explanations in a game-theoretic context. In particular, we provide the following contributions:

\begin{enumerate}[leftmargin=*,noitemsep,nolistsep]
\item We model the interactions between an ML system and an adversary as a \emph{two-player continuous-time signaling game}, where the variance of the generated explanations (by the ML system) evolve according to a \emph{stochastic differential equation (SDE)} (see Section \ref{sec:model}).  
\item We then characterize the \emph{Markov Perfect Equilibrium (MPE)} of the above stochastic game as a pair of two optimal functions $U(\pi)$ and $L(\pi)$, where $U(\pi)$ represents the optimal variance path for the explanations generated by the system, $L(\pi)$ represents the optimal variance path for the explanations given by the system to an adversary after adding some noise, and $\pi$ represents the belief of the system about the type of the adversary (see Section \ref{sec:model} and \ref{sec:analysis}).
\item We evaluate the game for different gradient-based explanation methods, namely, \emph{Integrated Gradients} \cite{sundararajan2017axiomatic}, \emph{Gradient*Input} \cite{shrikumar2016not}, \emph{LRP} \cite{bach2015pixel} and \emph{Guided Backpropagation} \cite{springenberg2014striving}.
We utilize five popular datasets, namely, Purchase, Texas, CIFAR-10, CIFAR-100, and Adult census dataset in the experiments. Then, we demonstrate that the capability of an adversary to launch MIA depends on different factors such as the chosen explanation method, input dimensionality, model size, and the number of training rounds (see Section \ref{sec:eval}).

\end{enumerate}

\section{Background and Preliminaries}
\label{sec:background}

%\subsection{Machine Learning} 
In this section, we provide a brief background of some important technical concepts, which will be useful in understanding our signaling game model later.
%\emph{two-player continuous-time Signaling Game}.

\subsection{Machine Learning}
\label{subsec:ml}
An ML algorithm (or model) is typically used to find underlying patterns within (vast amounts of) data, which enables systems that employ them to learn and improve from experience. In this work, we assume an ML model ($F:\mathbb{R}^{n} \to \mathbb{R}$) that performs a classification task, i.e., maps a given input vector $\overrightarrow{x}$ (with $n$ features) to a predicted label $y$. Given a labeled training dataset $\mathcal{X}_{tr}$ comprising of data points or vectors $\overrightarrow{x}$ and their corresponding labels $y$, any ML model $F$ is defined by a set of parameters $\theta$ taken from some parameter space $\Theta$ (and represented as $F_{\theta}$), and the model ``learns'' by calculating an optimal set of these parameters $\hat{\theta}$ on the training dataset using some optimization algorithm. 
%To be explicit, we may write $F_{\theta}(\overrightarrow{x})$ to refer to the model $F(\overrightarrow{x})$ with parameters $\theta$. 
In order to train ML models in such a supervised fashion, one needs to determine an optimal set of parameters $\hat{\theta}$ (over the entire parameter space) that empirically minimizes some loss function $l$ over the entire training data:
$$ \hat{\theta} \in \argmin_\theta \underset{(\overrightarrow{x},y)\in \mathcal{X}_{tr}} {\mathbb{E}} l(F_{\theta}(\overrightarrow{x}), y) $$
Here, the loss function $l(\cdot)$ intuitively measures how ``wrong'' the prediction $F_{\theta}(\overrightarrow{x})$ is compared to the true label $y$. One popular approach to train ML models is \emph{Stochastic Gradient Descent (or SGD)} \cite{bottou2010large} which solves the above problem by iteratively updating the parameters:
{
$$ \theta \gets \theta - \alpha \cdot \nabla_{\theta}(\sum_{(\overrightarrow{x},y)\in \mathcal{X}_{tr}} l (F_{\theta}(\overrightarrow{x}), y))$$
}
where $\nabla_{\theta}$ is the gradient of the loss with respect to the weights $\theta$ and $\alpha$ is the learning rate which controls by how much the weights $\theta$ should be changed.

\subsection{Gradient based Explanations} 
\label{subsec:gradient}
For some input data point $\overrightarrow{x} \in \mathbb{R}^n$ and a classification model $F_{\theta}$, an explanation method $\mathcal{H}$ simply explains model decisions, i.e., it outputs some justification/explanation of why the model $F_{\theta}$ returned a particular label $y=F_{\theta}(\overrightarrow{x})$. %In addition to the trained model $F_{\theta}$ and data point under question $\overrightarrow{x}$, some explanation methods may also employ additional information such as active access to model queries \cite{adler2018auditing} \cite{datta2016algorithmic} \cite{ribeiro2016should}, prior over the training data distribution \cite{baehrens2010explain} and information about the model such as its type or source code \cite{datta2017use} \cite{ribeiro2018anchors}. 
In this work, we consider feature-based explanations, where the output of the explanation function is an influence (or attribution) vector 
%$\mathcal{H}(\overrightarrow{x}) = \{\mathcal{H}_1(\overrightarrow{x}), \mathcal{H}_2(\overrightarrow{x}), ..., \mathcal{H}_n(\overrightarrow{x})\}$
and where the element $\mathcal{H}_i(\overrightarrow{x})$ of the vector represents the degree to which the $i^{th}$ feature influences the predicted label $y$ of the data point $\overrightarrow{x}$. One of the most commonly employed feature-based explanation method is the \textit{backpropagation-based} method, which as the name suggests, uses (a small number of) backpropagations  from the prediction vector back to the input features in order to compute the influence of each feature on the prediction. We next briefly describe some popular backpropagation-based explanation methods:\\
\textbf{Gradients: }In gradient-based explanations, $\mathcal{H}_i(\overrightarrow{x}) = \frac{\partial F_{\theta}}{\partial x_i}(\overrightarrow{x})$. Simonyan et al. \cite{simonyan2014deep} employed only absolute values of prediction vector gradients (during backpropagation) to explain predictions by image classification models, however negative values of these gradients are also useful (in other applications). We denote a gradient-based explanation function as $\mathcal{H}_{GRAD}(\overrightarrow{x})$. Shrikumar et al. \cite{shrikumar2016not} proposed to compute the \emph{Hadamard product} of the gradient with the input ($x_i \times \frac{\partial F_{\theta}}{\partial x_i}(\overrightarrow{x})$) to improve the numerical interpretability of released explanation. However, in our setting as the adversary already has the input vector $\overrightarrow{x}$, releasing this product is equivalent to releasing $\mathcal{H}_{GRAD}$, and so we do not consider it separately. \\
\textbf{Integrated Gradients: }Integrated gradients are obtained by accumulating gradients, computed at all points along the linear path from some baseline $\overrightarrow{x}^{'}$ (often $\overrightarrow{x}^{'} = \overrightarrow{0}$ ) to the actual input $\overrightarrow{x}$ \cite{sundararajan2017axiomatic}. In other words, integrated gradients are the path integral of the gradients along a straight-line path from the baseline $\overrightarrow{x}^{'}$ to the input $\overrightarrow{x}$. The integrated gradient (represented by us as $\mathcal{H}_{IGRAD}$) along the $i$-th feature for an input $\overrightarrow{x}$ and baseline $\overrightarrow{x}^{'}$ is defined as:
%\vspace{-0.4cm}
\begin{equation}
\mathcal{H}_{IGRAD}(\overrightarrow{x}_i) = (x_{i} - x_{i}^{'}) . 
\int_{\alpha=0}^{1} \frac{\partial F(\overrightarrow{x}^{'} + \alpha (\overrightarrow{x} - \overrightarrow{x}^{'}))}{\partial x_i}  \nonumber
\end{equation}

\noindent
\textbf{Layer-wise Relevance Propagation (LRP): }In this method \cite{bach2015pixel}, attributions are obtained by doing a backward pass on the model network. The algorithm defines the relevance in the last layer $L$ as the output of that layer itself, and for the previous layers, it redistributes the layer's relevance according to the weighted contribution of the neurons of the previous layer to the current layer's neurons. \\
\textbf{Guided Backpropagation: }Guided Backpropagation \cite{springenberg2014striving} is an explanation method designed for networks with positive ReLu activations. It is a different version of the gradient where only paths with positive weights and positive ReLu activations are taken into account during backpropagation. Hence, it only considers positive evidence for a specific prediction. While being designed for ReLu activations, it can also be used for networks with other activations.
\subsection{Membership Inference Attacks} 
\label{subsec:mia}
In membership inference attacks (MIA), the goal of an adversary, who is in possession of a target dataset $\mathcal{X}_{tgt} \subset \mathcal{R}^n$, is to determine which data points in this set belong to the training data set $\mathcal{X}_{tr}$ of some target model $F_{\theta}$, and which do not. This attack is accomplished by determining a function which for every point $\overrightarrow{x} \in \mathcal{X}_{tgt}$ is able to accurately predict if $\overrightarrow{x}$ was also present in $\mathcal{X}_{tr}$ or not. As trained ML models exhibit lower loss for members, compared to non-members, research in the literature \cite{long2017towards} has shown that an appropriately defined threshold (on the loss function) can be used to distinguish membership of target data points. As a matter of fact, Sablayrolles et al. \cite{sablayrolles2019white} showed that it is possible to determine an optimal threshold for such attacks under certain assumptions. However, it is easy to see that this attack (based on a pre-computed loss function threshold) will not work if the adversary does not have access to the true labels of the target data points and the loss function of the target model $F_{\theta}$.

In order to overcome this issue, Shokri et al. \cite{shokri2021privacy} recently generalized such threshold-based attacks to include the prediction and feature-based explanation vectors, which may be much more readily available compared to the data point labels and loss functions. In their proposed attack, they compute a threshold ($\tau_P$) for the variance of the prediction vector and similarly a threshold ($\tau_E$) for the variance of the feature-based explanation vector, and then use those thresholds to determine data point membership as follows:
%\vspace{-0.1cm}

{\scriptsize
\begin{align*}
    \mathrm{Membership}_{Pred,\tau_P}(\overrightarrow{x}) = 
    \begin{cases}
    \text{True},        & Var(F_{\theta}(\overrightarrow{x})) \geq \tau_P \\ \nonumber
    \text{False},    & otherwise
    \end{cases}
\end{align*}
}
%\[
{\scriptsize
\begin{align*}
    \mathrm{Membership}_{Expl,\tau_E}(\overrightarrow{x}) = 
    \begin{cases}
    \text{True},        & Var(\mathcal{H}_{GRAD}(\overrightarrow{x})) \leq \tau_E \\
    \text{False},    & otherwise
    \end{cases}
\end{align*}
}
%\]
where the variance of some vector $\overrightarrow{v} \in \mathbb{R}^n$ is calculated as:
$$Var(\overrightarrow{v}) = \sum_{i=1}^{n} (v_{i} - \mu_{\overrightarrow{v}})^{2} \text{,} \quad \text{where} \mu_{\overrightarrow{v}} = \frac{1}{n} \sum_{i=1}^{n} v_{i}$$
For the former case, the intuition is that a low model loss typically translates to a prediction vector that is dominated by the true label, resulting in a high variance, which may be indicative of model certainty, and thus, the data point (under consideration) as being a member of the training dataset. For the latter case (threshold attack based on feature-based explanation variance), the motivation is similar. In this work, since our focus is on threshold-based membership inference attacks, thus, the crucial question is: \emph{how to compute or determine the discriminating threshold, i.e., $\tau_E$}? 
%For points closer to the decision boundary, and thus unlikely being part of the training set, smaller changes in a particular feature may result in significant changes in the model's prediction, indicating a large explanation vector variance. In other words, a small explanation vector variance is indicative of the point being a part of the training dataset. However, a central question in both the above threshold-based membership inference attacks is, \emph{how to compute or determine the discriminating threshold  ($\tau_P$ or $\tau_E$)}?

%Shokri et al. \cite{shokri2021privacy} discuss two potential strategies to compute this threshold. First strategy is to trivially compute the threshold for a set of data points, comprising both members and non-members, provided that their membership is already known. Although such a strategy results in an optimal threshold providing the best possible accuracy to the adversary, it is not very meaningful to assume that data point membership is known in advance! The second strategy assumes that the adversary possesses some labeled data points from the target distribution, and uses it to train a small number of shadow/reference models. The variance thresholds are then computed for these references models, and then used to carry out membership inference of some target data points. As we discussed earlier, this strategy makes an important assumption that the adversary knows the target model type and its training hyper parameters.

\subsection{Geometric Brownian Motion}
\label{subsec:gbm}
As discussed earlier, the goal of an adversary is to reach some expected variance threshold in order to launch explanation-based threshold attacks. We assume that the adversary will try to accomplish this by repeatedly interacting with the ML model, using appropriate queries/data points from the target model's input space and any available historical interaction information. The pattern of the feature-based explanation (vector) variance in these interactions is expected to follow an increasing path, with both positive and negative periodic and random shocks (or fluctuations). As a result, we represent the evolving explanation variance (denoted as $EX^{v}$) due to the adversary's repeated interactions with the ML model as a continuous-time stochastic process that takes non-negative values, specifically as a \emph{Geometric Brownian Motion (GBM)} process. 
%The \texttt{system} uses an explanation method $\mathcal{H}$ to generate explanations for labels predicted by the machine learning model $F_{\theta}(\overrightarrow{x})$. We assume that there are random fluctuations in the generated explanations variance as mentioned in the introduction.In summary, we assume that the explanation variance ($EX^{v}$) is a stochastic process that follows a GBM, and which can experience shocks both upward and downward.  
A GBM is a generic state process $s_t$ that satisfies the following stochastic differential equation (SDE):
\begin{equation}
ds_{t} = a(s_t,u(s_t,t),t) s_{t} dt + b(s_t,u(s_t,t),t) s_{t} dW_{t} \nonumber
\end{equation}
where, $a(s_t,u(s_t,t),t)$ and $b(s_t,u(s_t,t),t)$ are the drift and volatility parameters of the state process $s_t$, respectively, $W_{t}$ is a standard Brownian motion with mean $=$ $0$ and variance $=$ $t$, and $u(s_t,t)$ is the control.

\subsection{Optimal Control and the Stopping Problem} 
\label{subsec:ctbw}
In this work, we are trying to model repeated interaction between two agents, the adversary and the ML model, where each agent makes an optimal control decision in each interaction instance, i.e., either to continue or stop the process of sending queries and (their) explanations, respectively. Such problems involving optimal control are usually modeled with \emph{Bellman's} equation and solved using optimization techniques such as \emph{Dynamic Programming}. Below, we present a very generic description of modeling using this technique. Without loss of generality, let's assume a system with two agents and let $s_t$ represent the system state at time $t$. Let $u_i(s_t,t)$ represent the control of agent $i$ when the system is in state $s_t$ at time $t$. The value function, denoted by $H_i(s_t,t)$, represents the optimal payoff/reward of the agent $i$ over the interval $t=[0, T]$ when started at time $t=0$ in some initial state $s_0$, and can be written as:
%. Now, given that $s_{t}$ follows a GBM, the value function ($H(s_t,t)$) 
$$H_i(s_t, t) = \max_{u_i}\int_{0}^{T} f(s_t,u(s_t,t),t)dt $$ 
where, $f(s_t,u(s_t,t),t)$ is the instantaneous payoff/reward a player can get given the state ($s_t$) and the control used ($u$) at time $t$. For continuous-time optimization problems, the Bellman equation is a \emph{partial differential equation or PDE}, referred to as the \emph{Hamilton Jacobi Bellman (HJB)} equation, and can be written as:
{
$$rH(s_t, t) = f(s_t,u^{*},t) + \frac{\partial H}{\partial t} + \frac{\partial H}{\partial s_t}a(s_t,u^{*},t) $$
$$+ \frac{1}{2}\frac{\partial^{2} H}{\partial s_t^{2}}b(s_t,u^{*},t)^{2}$$
$$u^{*} = u(s_t,t) = \text{optimal value of control variable}$$
}
As outlined later, we represent the value functions of both the agents in this work (i.e., the adversary and the ML model) using the above equation. The optimal control $u$ for an agent is a binary decision, with $u=1$ representing ``stopping" the task being done in the previous time instant, while $u=0$ representing continuation of the task from the previous time instant.

\noindent \textbf{Stopping Problem:}
A \emph{stopping problem} models a situation where an agent must decide whether to continue the activity he/she is involved in (in the current time instant) and get an instantaneous flow payoff, $f(s_t,u(s_t,t),t)$, or cease it and get the termination payoff, $\lambda(s_t, T)$. It is determined based on the payoff he/she is expected to receive in the next instant. 
%In the case of two interacting agents (as assumed in this paper), each needs to make an optimal decision on whether to continue their respective activities and get an instantaneous flow payoff, $f(s_t,u(s_t,t),t)$ or stop the activity they were involved in (previous time instant) and get the termination payoff, $\lambda(s_t, T)$. 
If $s^{*}_{t}$ is the state boundary value at which an agent decides to stop and get the termination payoff, then the solution to the stopping problem is a stopping rule:
    \[
    u(s_t,t) = 
    \begin{cases}
    \text{stop},        & s_{t} >= s^{*}_{t}  \\
    \text{continue},    & s_{t} < s^{*}_{t}
    \end{cases}
    \]
 In other words, when the agent decides to stop, he/she gets: 
    {
    $$H(s_t,T) = \lambda(s_t, T)\quad \forall s_t \geq s^{*}_{t}$$
    }

\noindent
\textbf{Value Matching and Smooth Pasting Conditions:}
In order to solve the HJB equation outlined above, two boundary conditions are required, which we describe next. The first condition, called the \emph{value matching condition}, defines a constraint at the boundary which tells an agent that if they decide to stop (at that defined boundary), then the payoff it would get is at least the same as the previous time instant. The value matching condition determines whether the value function is continuous at the boundary or not: 
{
    $$H(s^{*}_{t},t) = \lambda(s^{*}_{t}, t) \quad \forall t$$
}
However, as the boundary $s^{*}_{t}$ is also an unknown variable, we need another condition which will help in finding $s^{*}_{t}$ along with $H(s_{t},t)$. 
The \emph{smooth pasting condition} helps in pinning the optimal decision boundary, $s^{*}_{t}$. Intuitively, it also helps to formulate an agent's indifference between continuation and stopping. 
{
$$H_{s_t}(s^{*}_{t},t) = \lambda_{s_t}(s^{*}_{t}, t) \quad \forall t$$
}
where $H_{s_t}(s^{*}_{t},t)$ is the derivative of $H(s^{*}_{t},t)$ with respect to the state $s_t$. If one or both the above conditions are not satisfied, then stopping at the boundary $s^{*}_{t}$ can't be optimal. Therefore, an agent should continue and again decide at the next time instant.

\section{Game Model}
\label{sec:model}

%"Optimal adversary's query choices will have the property that the explanation variance will appear like GBM." 

%"Impact of variance on the query formulation"

%"How and adversary is getting the explanations. What's the impact of variance on the query space"

Next, we present an intuitive description of the problem followed by its formal setup as a signaling game. Further, we also characterize the equilibrium concept in this setup.
\subsection{Intuition}
\label{subsec:intuition}
We consider a platform, referred to as the \texttt{system}, that makes available an ML model and a feature-based explanation or attribution method as a (black box) service. Customers of this platform, referred to as \texttt{end-users}, seek a label and its explanation for their query (sent to the \texttt{system}), but cannot download the model itself from the platform. 
%A query is classified as \emph{suspicious} if the explanation generated for its label lies near some boundary of maximum relevant explanations which can be given by the \texttt{system} to the \texttt{end-user}, otherwise, it is \emph{non-suspicious}. The \texttt{system} can serve two types of \texttt{end-users}: \emph{honest} and \emph{malicious}. Honest \texttt{end-users} have no malicious goal (of compromising the \texttt{system}) and always send non-suspicious queries, 
%and queries sent by honest \texttt{end-users} to the \texttt{system} are called referred to as ``\emph{signals}" (or ``\emph{signaling}"). 
The \texttt{system} can serve two \emph{types} of \texttt{end-users}: \emph{honest} and \emph{malicious}, but it is unaware of the type of \texttt{end-user} it is interacting with. An honest \texttt{end-user}'s goal is to formulate a query (from the ML model's input space) and get an explanation (as outlined in section \ref{subsec:gradient}) for the label generated by the model. However, a malicious \texttt{end-user}'s primary goal is to carry out an explanation threshold-based MIA against the ML model by employing the variance of the (gradient-based) explanations received from the \texttt{system} without getting detected by the \texttt{system}.
%To accomplish this objective without getting detected (or blocked) by the \texttt{system}, a malicious \texttt{end-user} will attempt to frame queries by \emph{mimicking} honest \texttt{end-users} as much as possible to confuse the \texttt{system}. 

\changed{The malicious \texttt{end-user} attempts to accomplish this by repeatedly interacting with the \texttt{system} to get explanations for the set of (input) query space that an adversary formulates, utilizing the prior variance history (which we have modeled using GBM). As explanation-based MIAs \cite{shokri2021privacy} utilize explanation variance as a threshold, we believe it is appropriate to model this variance as a GBM process. The feature-based explanation (vector) variance pattern in these interactions is expected to follow an increasing path, with both positive and negative periodic and random shocks (or fluctuations). Hence, the modeling explanation variance as GBM helps to integrate historical data, and Brownian motion ensures that the explanation variance remains positive. 
%\commentM{I believe this is still not very strong as to why variance is a GBM? Can we add some more motivation?}
%, and not the query space. 
In other words, we assume that the optimal adversary's query choices will have the property such that the resulting explanation variance will follow a GBM. Note: When we refer to modeling or formulating the query space, it is intended to highlight that an adversary resolves this task using the modeled explanation variance, which serves as the central focus of our paper. Also, our goal is to provide a sound mathematical formulation of the existence of the explanation variance threshold that an adversary can utilize to launch MIAs. Thus, we are not concerned with how an adversary models the query space - our focus is to mathematically prove the existence of such an optimal explanation variance threshold.} The malicious \texttt{end-user} must strategically decide to stop eventually, at an appropriate moment (state), in order to accomplish the attack objective. We model this behavior as a continuous time signaling game. If malicious \texttt{end-user} behavior is not detected by the \texttt{system}, and is deemed as honest, then this behavior is referred as \emph{pooling} or \emph{on-equilibrium path} behavior. Alternatively, if the malicious \texttt{end-user} deviates from this behavior, it is referred to as the \emph{separating} or \emph{off-equilibrium} path behavior. In summary, as the game evolves, the malicious \texttt{end-user} attempts to carry out the threshold-based MIA by utilizing the information (labels+explanations) accumulated up until that point.

%They accomplish this by imitating the honest \texttt{end-users} (until the time is right), and then carry out the attack with the information (labels+explanations) accumulated up until that point \mj{I have removed the part about honest and malicious queries. If I understand correctly, there are not honest or malicious queries - only signaling and not (stop) signaling}.
%In other words, an honest \texttt{end-user} is always expected to signal, while the malicious \texttt{end-user} will decide whether to signal or stop signaling and carry out the privacy attack. 
%Queries are classified as honest or malicious based on the cost of query formulation by the \texttt{end-user}. 
%MJ- put the below sentence in assumptions section
%We assume that the malicious \texttt{end-user} has enough resources to compromise the \texttt{system}. Thus, 
More specifically, the malicious \texttt{end-user} must decide (at each interaction instance) to either continue getting explanations by formulating (input) queries based on past behavior or stop the play and finally attack the \texttt{system}, as described in detail later in Section \ref{subsec:eqdesc}. 
%Imitating the honest \texttt{end-user} (or signaling) is costly for a malicious \texttt{end-user} due to the associated costs (e.g., additional query framing and processing), and thus he/she should optimally decide to either continue getting explanations through honest queries from the \texttt{system} or to deviate from honest behavior (signaling) to compromise the \texttt{system}. 
This is important as the malicious \texttt{end-user} wants to avoid being detected by the \texttt{system} prior to compromising it.
% (after sending a series of queries to the system) by either inferring the membership of some data points or by creating targeted or non-targeted adversarial attacks on explanations, etc. based on an optimal variance path $U(\pi)$ (outlined in detail later in Section \ref{subsec:eqdesc})
The \texttt{system}, on the other hand, on receiving requests from an \texttt{end-user} must decide whether to continue giving explanations and how much noise/perturbation to add to it, or to just block the \texttt{end-user} based on an optimal variance path $U(\pi)$ (outlined in detail later in Section \ref{subsec:eqdesc}). It should be noted that the \texttt{system} has \emph{imperfect information} about the type of the \texttt{end-user} and this information is only contained in its Bayesian prior or belief ($\pi$). Thus, the added noise/perturbation to the generated explanation is based on the \texttt{system}'s belief pertinent to the activity history of the \texttt{end-user}. 
%\mj{We need to clarify the part about honest and malicious queries. What is a malicious query? After that MJ, please complete this subsection.}
%, i.e., queries whose labels and related explanations cannot be collectively used to compromise the ML model's privacy.
%Given the above model, our focus is to observe the pooling and semi-pooling outcomes of the game. Pooling here means that both types of \texttt{end-users} send the same type of query. In our model, pooling means both types of \texttt{end-users} would send the honest signal (since we have assumed honest user always signals). Semi-pooling means that while \texttt{honest users} send honest queries, \texttt{malicious users} will mix between sending honest and malicious queries/signals. 
%\kavita{I think it is clear from this statement that we are considering the same ens-user for the whole play} 

\changed{Based on this stopping game formulation, we structure the model payoffs for both the \texttt{system} and the \texttt{end-users} (malicious and honest). Additionally, according to each interaction instance between the \texttt{system} and the \texttt{end-user}, we formulate the noise and the stopping responses. As the \texttt{system} has imperfect knowledge about the type of \texttt{end-user} it interacts with, the added noise/perturbation to the generated explanation is based on the system’s belief pertinent to the activity history of the end-user. For an honest \texttt{end-user}, the value of an explanation lies in its relevance — the information it contains explaining the model's decision for the query sent to the benign \texttt{end-user}. This relevance inherently can be explained by the variance of the explanation, varying accordingly. Conversely, for a malicious \texttt{end-user}, the value of an explanation also depends on its relevance — the amount of information it holds that can be exploited by the malicious \texttt{end-user} to launch MIAs. A detailed explanation of the payoffs and its design is outlined in Section \ref{subsec:setup}.
}

In this preliminary effort, we formally model the above interactions between a single \texttt{end-user} (type determined by nature) and the \texttt{system} within a stochastic game-theoretic framework, and further analyze it to answer the following two high-level questions: When does a malicious \texttt{end-user} decide to stop the play and finally compromise the \texttt{system}? How does the \texttt{system} make the strategic decision to block a potential malicious \texttt{end-user} while continuing to give relevant explanations to potentially honest \texttt{end-users}? 
%Our primary insight is that, as the explanations evolve in a stochastic environment, the incentive to signal disappears at some point in time, i.e. \texttt{malicious end-user} will stop signaling or deviate from the honest behavior and try to compromise the \texttt{system}. Therefore, the critical question (from the \texttt{malicious user's} perspective) our analysis answers is that at what explanation threshold he/she should decide to deviate from honest behavior. We are assuming that when the game starts, both types of \texttt{end-user} signals i.e. pooling behavior is assumed. The \texttt{system} will want to block the \texttt{malicious end-user} if, after a series of queries sent by the \texttt{malicious user}, the generated explanation is at least some upper threshold, $u_{th}$. $u_{th}$ is the maximum relevance of the explanation generated by an explanation method. On the other hand, if the explanation given to the \texttt{end-user} gets equal to some lower threshold, $l_{th}$, then there is little threat from the \texttt{system} in the present moment. $l_{th}$ is the maximum relevant explanation given to the \texttt{end-user} by the \texttt{system} (\texttt{system} having a belief of 1 i.e. knows with certainty that \texttt{end-user} is of honest type). Hence, the \texttt{end-user} of malicious type no longer wants to send costly (honest) signals and, consequently, has an incentive to deviate from the honest behavior and attack the \texttt{system} to achieve his/her goal. This effectively ends the game.

\subsection{Setup and Assumptions}
\label{subsec:setup}
We model the above scenario as a two-player, continuous-time, imperfect-information game with repeated play. We opted for the continuous-time framework because a malicious end-user can choose to deviate (stopping time) from the pooling behavior at any time, or a system can decide to block the end-user at any time. \changed{Moreover, since we are using GBM to model explanation variance and it can have abrupt transitions, continuous-time modeling offers a more subtle representation of its evolution. In literature, problems built upon the stopping time are also modeled in the continuous-time framework for the same reason.} The game has two players: Player 1 is the \texttt{end-user}, of privately known type $\Theta \to \{h,m\}$ (i.e., honest or malicious), who wants to convince Player 2 (i.e., \texttt{system}) that he/she is honest. The game begins with \emph{nature} picking an \texttt{end-user} of a particular type, and we analyze repeated play between this \texttt{end-user} (selected by nature) and the \texttt{system}, which occurs in each continuous-time, $t \in R$. %\commentK{There was a comment by reviewer C who rejected the paper stating "Nature picks (and therefore the defender models) a benign user with the same probability as a malicious one which wildly over-estimates adversarial probability expected in the real world. ". Not sure how to address this comment.} \commentM{Can we say that nature is biased and picks with a probability $\Theta$ that matches the occurrence of malicious users in practice/real-world? I am not sure if this assumption has any significant impact on the model. The system still has probabilistic (imperfect) knowledge about nature's pick!}
%\kavita{I can remove t as well as I am using it as the subscript of many variables. So, I have stated that each $t$ belongs to $R$}. 
As the \texttt{system} has imperfect information about the type of \texttt{end-user}, it assigns an initial belief $\pi_{0}= Pr(\Theta = h)$.
% that the received query was from an \texttt{honest user}. 
We assume both players are \emph{risk neutral}, i.e., indifferent to taking a risk, and each player \emph{discounts} payoffs at a constant rate $r$. 
%We have assumed that the explanation generated for a specific label of the query sent by the \texttt{end-user} is subjected to persistent shocks (explanation varies as it depends on the corresponding features contribution) as explanation is being generated using an explanation method, thus varies from query to query. As a result, we have modeled the generated explanation as a stochastic process that follows a Geometric Brownian motion. We allow this process to have an upward or downward drift. Hence, our results apply equally to scenarios where the relevance of an explanation given to the \texttt{end-user} either decreases or increases. 
%As detailed in \ref{subsec:gbm}, the equation of motion of the 
Variance ($EX_{t}^{v}$) computed for an explanation generated by an explanation method of the \texttt{system} follows a GBM, and is given by:

$$dEX_{t}^{v}=\mu EX_{t}^{v} dt + \sigma EX_{t}^{v} dW_{t}$$

where, $\mu$ is the constant \emph{drift} and $\sigma > 0$ is the constant \emph{volatility} of the variance process $EX_{t}^{v}$, and $EX_{0}^{v} = ex_{0}^{v} > 0$. 
%Also, $\mu$  and $\sigma$ are identified as constants in the game. 
$W_{t}$ is a standard Brownian motion with mean = $0$ and variance = $t$. To ensure finite payoffs at each continuous time $t$, we assume $\mu < r$. The state of the game is represented by the process ($EX_{t}^{v}, \pi_{t}$), where
%, where $ex_{t}^{v}$ is the realization of the process $EX_{t}^{v}$ 
$\pi_{t}$ is the belief of the \texttt{system} about the type of the \texttt{end-user} at time $t$.
%, as these are the two parameters based on which both players formulate their strategy.

%\noindent
%\textit{Assumption 2: } 
%As discussed before, an \texttt{end-user} sends a query to the \texttt{system} to get a label and it's explanation. In response, t
The \texttt{system} wants to give informative or relevant explanations to the honest \texttt{end-user}, but noisy explanations to the malicious \texttt{end-user}. Hence, depending on the \texttt{system}'s belief about the type of the \texttt{end-user}, it will decide how much noise/perturbation to add to each released explanation, according to the generated variance. Let $U(\pi_{t})$ denote the optimal variance path (or functional path) for the \texttt{system} - a non-increasing cut-off function which tells the \texttt{system} the optimal explanation variance computed for an explanation generated by an explanation method and $L(\pi_{t})$ denote the optimal explanation variance path for the \texttt{end-user} - an increasing cut-off function which tells the \texttt{end-user} the optimal explanation variance for the explanation given by the \texttt{system} at given belief $\pi_{t}$. To simplify the resulting analysis, we assume that the explanations variance computed by the \texttt{system} and explanations variance computed by the \texttt{end-user} are just different realizations of the explanation variance process $EX_{t}^{v}$. We denote $ex^{v}_{s_{y}, t}$ and $ex^{v}_{e_{u}, t}$ as the \texttt{system}'s and \texttt{end-user}'s realization of the process $EX_{t}^{v}$, respectively. Moreover, as the \texttt{system} would add some calculated noise to the generated explanation based on its Bayesian belief, we assume that $U(\pi_{t}) \geq L(\pi_{t}),\  \forall \pi_{t}$.
%, which can experience shocks both upward and downward.\\
%At any time $t \geq 0$, the \texttt{system} (or player 2) can terminate the game by choosing to suspend or block the account of the \texttt{end-user} (or player 1). Prior to block by the \texttt{system} (player 2), the \texttt{end-user} (or player 1) determines whether to send an honest query or a malicious query. If either type of \texttt{end-user} displays honest behavior, we say \texttt{end-user} is signaling. Therefore, type $h$ will always signal and type $m$ will decide when to stop signaling i.e. deviate from the honest behavior. \\
%\noindent
%\textit{Assumption 3: }
Finally, as we are only interested in modeling the interactions between a malicious \texttt{end-user} and the \texttt{system}, any reference to an \texttt{end-user} from this point on implies a malicious \texttt{end-user}, unless explicitly stated otherwise.
%\noindent
%\textit{Assumption 4: } In our modeling scenario, the \texttt{malicious end-user's} eventual objective is to compromise the \texttt{system}, i.e., compromise the privacy of the \texttt{system's} machine learning model by employing the information available from the received explanations. To accomplish this objective without getting blocked (or caught) by the \texttt{system}, the \texttt{malicious end-user} will attempt to frame queries by mimicking the \texttt{honest user} (queries) as much as possible to confuse the \texttt{system}. Eventually, when the time is right, the \texttt{malicious end-user} strategically deviates from this signaling strategy (i.e., mimicking the \texttt{honest user}) to accomplish the attack objective. This deviation is accomplished by the \texttt{malicious end-user} selecting queries from a different, much larger, query space than that of the \texttt{honest user}. In summary, as the game evolves, a \texttt{malicious user} can choose to signal (mimic the honest type) by sending queries similar to that of \texttt{honest user} or deviate from signaling by reverting to its own query space. \\
% In the forthcoming analysis, we will attempt to show the conditions under which a unique MPE exists in our signaling game model of this problem. 
Next, we outline a few other relevant model parameters before characterizing the concept of equilibrium in the proposed game model.
% and variables for both the players of our game model, i.e., the \texttt{system} and the \texttt{end-user}, which will be required for this analysis. \\
%\noindent
%\begin{itemize}[leftmargin=*]

\noindent \textbf{Information Environment:} Let $\mathcal{F}_{t} = \sigma(\{EX_{s}^{v}\}: 0 \leq s \leq t)$ be the \texttt{end-user}'s information environment, which is the \emph{sigma-algebra} generated by the variance process $EX^{v}$. In other words, $\mathcal{F}_{t}$ represents the information contained in the public history of the explanation variance process. The \texttt{system}'s information environment is denoted by $\mathcal{F}_{t}^{+} = \sigma(\{EX_{s}^{v}, \phi_{s}\} : 0 \leq s \leq t)$, where $EX_{s}^{v}$ is the variance process representing the history of explanations variance and $\phi_{s}$ is the stochastic process representing the historical activity of the \texttt{end-user}. If $\rho$ is the time that \texttt{end-user} decides to stop, then $\phi_{t}$ = $\rho$ if $\rho \leq t$ and $\infty$ otherwise.

\noindent \textbf{Strategies:} Next, let us outline the strategy space for both the \texttt{end-user} and the \texttt{system}.
%We allow the \texttt{end-user} to play randomized strategy and the \texttt{system} to play pure strategies. The reason being \texttt{malicious end-user} decides to either signal or deviate in each continuous time $t$ till he/she has reached $l_{th}$, thus he/she plays randomized strategy. On the other hand, \texttt{system} optimally decides it's strategy based on it's computed belief about the type of \texttt{end-user}, thus plays pure strategy. 
%Fundamentally, pure strategies are (collections of) stopping times, one for each continuation time $t_{0} \geq 0$.

\begin{itemize}[leftmargin=*]
    \item \texttt{end-user}: We only define strategies for the malicious (type $m$) \texttt{end-user} as we have considered that only malicious \texttt{end-user} has an incentive to launch explanation-based MIA. We assume that the (malicious) \texttt{end-user} plays a \emph{randomized} strategy, i.e., at each time $t$, he/she either continues to interact with the \texttt{system} or stops querying and attacks the \texttt{system}. The \texttt{end-user}'s strategy space is dependent on the history of the variance for the explanations given by the \texttt{system}; hence it is a collection of $\mathcal{F}_{t}$ - adapted stopping times \{$\phi_{t}$\} such that $\phi_{t}$ = $\rho$ if $\rho \leq t$ and $\infty$ otherwise. 

    \item \texttt{system}: We assume \texttt{system} plays a pure stopping time ($\tau^{t}$) strategy, i.e., the time at which it blocks the \texttt{end-user} (if deemed malicious). \changed{We assumed this because the \texttt{system} aims to block the malicious end-user. Thus, continuous interaction with the \texttt{end-user} to provide model predictions and their labels is an implicit action for the \texttt{system}.} The \texttt{system}'s strategy is dependent on the evolution of the explanation variance process $EX^{v}$ and the record of \texttt{end-user}'s querying activity. Hence, the strategy space of the \texttt{system} is a collection of $\mathcal{F}_{t}^{+}$ - adapted stopping times $\{\tau^{t}\}$.
\end{itemize}

\noindent
We use a \emph{path-wise Cumulative Distribution Function (CDF)}, represented as $R_{t}^{t_{0}}$, to characterize how fast the computed variance at a given time $t$ is trying to reach the variance threshold (defined later). We compute this CDF from the probability density function ($p_{t}(ex^{v}_{s_{y}, t})$) of the GBM, given by:

%\vspace{-0.1cm}
{\scriptsize
\begin{equation}
    p_{t}(ex^{v}_{s_{y}, t}) = \frac{1}{\sqrt{2 \pi t} \sigma ex^{v}_{s_{y}, t}} \exp{\left(- \frac{[ln(ex^{v}_{s_{y}, t}) - (\mu - \sigma^{2}) t]^{2}}{2 \sigma^{2} t}\right)} \text{,} \nonumber
\end{equation}
}
where, $ex^{v}_{s_{y}, t} \in (0, \infty)$. In other words, the CDF ($R_{t}^{t_{0}}$) will give the probability of how close the computed explanation variance is to the explanation variance threshold at time $t$ starting from the explanation variance computed at time $t_{0}$, i.e., $ex^{v}_{s_{y}, 0}$.

\noindent \textbf{Beliefs:} Given information $\mathcal{F}_{t}^{+}$, the \texttt{system} updates its beliefs at time $t$ from time $t_{0} < t$ using \emph{Bayes' rule} shown below. It is defined as the ratio of the probability of the honest \texttt{end-user} sending queries to the \texttt{system} (set to 1) to the total probability of honest \texttt{end-user} and malicious \texttt{end-user} sending queries to the \texttt{system}. 
\[
    \pi_{t}= 
\begin{cases}
    \frac{1}{1 + (1-\pi_{t_{0}})R_{t}^{t_{0}}},        & \text{if} \quad \pi_{t_{0}} > 0 \quad \text{and} \quad \rho > t.  \quad \ \ (\textbf{i})\\
    0,                         & \text{if} \quad \rho \leq t \quad \text{or} \quad \pi_{t_{0}} = 0. \qquad (\textbf{ii})
\end{cases}
\]

Bayes' rule (i) is used when the \texttt{end-user} has not stopped communicating with the \texttt{system} ($\rho \geq t$) and the initial belief of the \texttt{system} about the \texttt{end-user}'s type is also not zero. 
%Thus, in this case \texttt{system} will compute it's belief using Bayes' rule at time $t$.
However, if the \texttt{system} has already identified the \texttt{end-user}'s type as $m$ or the \texttt{end-user} has already stopped communicating with the \texttt{system} and gets detected by it, then \texttt{system}'s belief $\pi_{t}$ will be zero, as indicated in (ii). \\

%\textbf{Equilibrium:} We employ the equilibrium concept of MPE in this work, where the state variable is the pair $(EX,\pi)$ consisting of the current explanation value and the current belief about the \texttt{end-user's} type. At all times, continuation behavior must be optimal and beliefs must be consistent with the Bayes' rule whenever possible. The Markovity conditions ensure that player's strategies depend only on the state variable pair $(EX,\pi)$. 

\begin{table}
\caption{Flow Payoff Coefficients}
%before and after Detection by the \texttt{system}.}
\label{tab:payoffs}
\centering
\resizebox{\columnwidth}{!}{%
\begin{tabular}{|c|c|c|c|}
\hline
\rowcolor{gray}
& \multicolumn{2}{|c|}{Before Detection} & After Detection \\ 
  & Pooling & Separation Starts & Detection and Block (Game Ends) \\
  \hline
 \texttt{end-user}, type $m$ & $P$ & $M_{NS}^{m}$ & $-k$ \\ 
 \hline
\texttt{end-user}, type $h$ & $P$ & $P$ & $P$ \\
\hline
 \texttt{system} & $r_{e}$ & $D_{NS}^{\Theta}$ & $D_{B}^{\Theta}$ \\
 \hline
\end{tabular}
}
\end{table}
%\lipsum[2-10]%
\noindent \textbf{Payoffs:} Table \ref{tab:payoffs} summarizes the flow payoff coefficients assumed in our game model. The \texttt{system} earns a reward of $D_{B}^{\Theta = m} = k EX_{t}^{v}$ for detecting and 
%observes an attempt by the \texttt{end-user} to compromise it using the computed belief, it  for successfully 
blocking the malicious \texttt{end-user}.
% as shown in table \ref{tab:payoffs}. 
The \texttt{end-user}'s type (malicious) is immediately revealed at this time, thus a cost of $-k$ is incurred by the \texttt{end-user}. In case of an interaction with an honest \texttt{end-user}, the \texttt{system} will always earn a payoff of $r_{e} EX_{t}^{v}$, i.e., $D_{NS}^{\Theta = h} = r_{e} EX_{t}^{v}$ and $D_{B}^{\Theta = h} = r_{e} EX_{t}^{v}$, while the honest \texttt{end-user} always earns a reward of $P EX_{t}^{v}$ in each stage of the game.
% as \texttt{honest user} doesn't have any incentive to attack the \texttt{system}. 
In case of a malicious \texttt{end-user} who keeps communicating with the \texttt{system} without being detected i.e., pools with the honest \texttt{end-user}, he/she receives a payoff (relevant explanation variance information) of $P EX_{t}^{v}$.
% is the relevant explanation given to the \texttt{end-user} who is considered honest by the \texttt{system}, i.e., the \texttt{end-user} who signals.
Prior to detection, if the malicious \texttt{end-user} stops and is able to compromise the \texttt{system}, then the \texttt{system} will have to pay a lump-sum cost of $D_{NS}^{\Theta=m} = -d^{'}$ and the malicious \texttt{end-user} will earn $M_{NS}^{m}= (M^{m}+d^{'}) EX_{t}^{v}$, where $M^{m} EX_{t}^{v}$ is the gain which relates to the explanation variance information gained from the \texttt{system}, $d^{'} EX_{t}^{v}$ is the benefit (can be monetary) achieved after attacking the \texttt{system}. Malicious \texttt{end-user} will also incur cost of deviation $d$. We make the following assumptions about the payoff coefficients: We assume that $D_{B}^{\Theta = m} = k EX_{t}^{v} > r_{e} EX_{t}^{v}$, as the \texttt{system} will gain more in successfully preventing the attack from the malicious \texttt{end-user}. When the malicious \texttt{end-user} decides to stop and attack the \texttt{system} and is not successful in compromising the \texttt{system}, then $PEX_{t}^{v} \geq M^{m}EX_{t}^{v}$ ($d^{'}=0$) as the \texttt{system} has not yet blocked the malicious \texttt{end-user} and because of the cost of deviation. 
%\end{itemize}

\subsection{Equilibrium Description}
\label{subsec:eqdesc}
%\textbf{Equilibrium:}  At all times, continuation behavior must be optimal and beliefs must be consistent with the Bayes' rule whenever possible. The Markovity conditions ensure that player's strategies depend only on the state variable pair $(EX,\pi)$.
%We employ the concept of Markov Perfect Equilibrium or MPE in this work, where the state variable is $(EX,\pi)$, consisting of the current explanation value and the current belief about the \texttt{end-user}'s type.
%In this subsection, we provide an intuitive description of the equilibrium concept and equilibrium characterization. 
A \emph{Markov Perfect Equilibrium (MPE)} consists of a strategy profile and a state process $(EX^{v},\pi)$ such that the malicious \texttt{end-user} and the \texttt{system} are acting optimally, and $\pi_{t}$ is consistent with Bayes' rule whenever possible (in addition to the requirement that strategies be Markovian). A unique MPE occurs when the two types of \texttt{end-users} display \emph{pooling} behavior.
\begin{figure}
\centering
\includegraphics[width=0.65\linewidth]{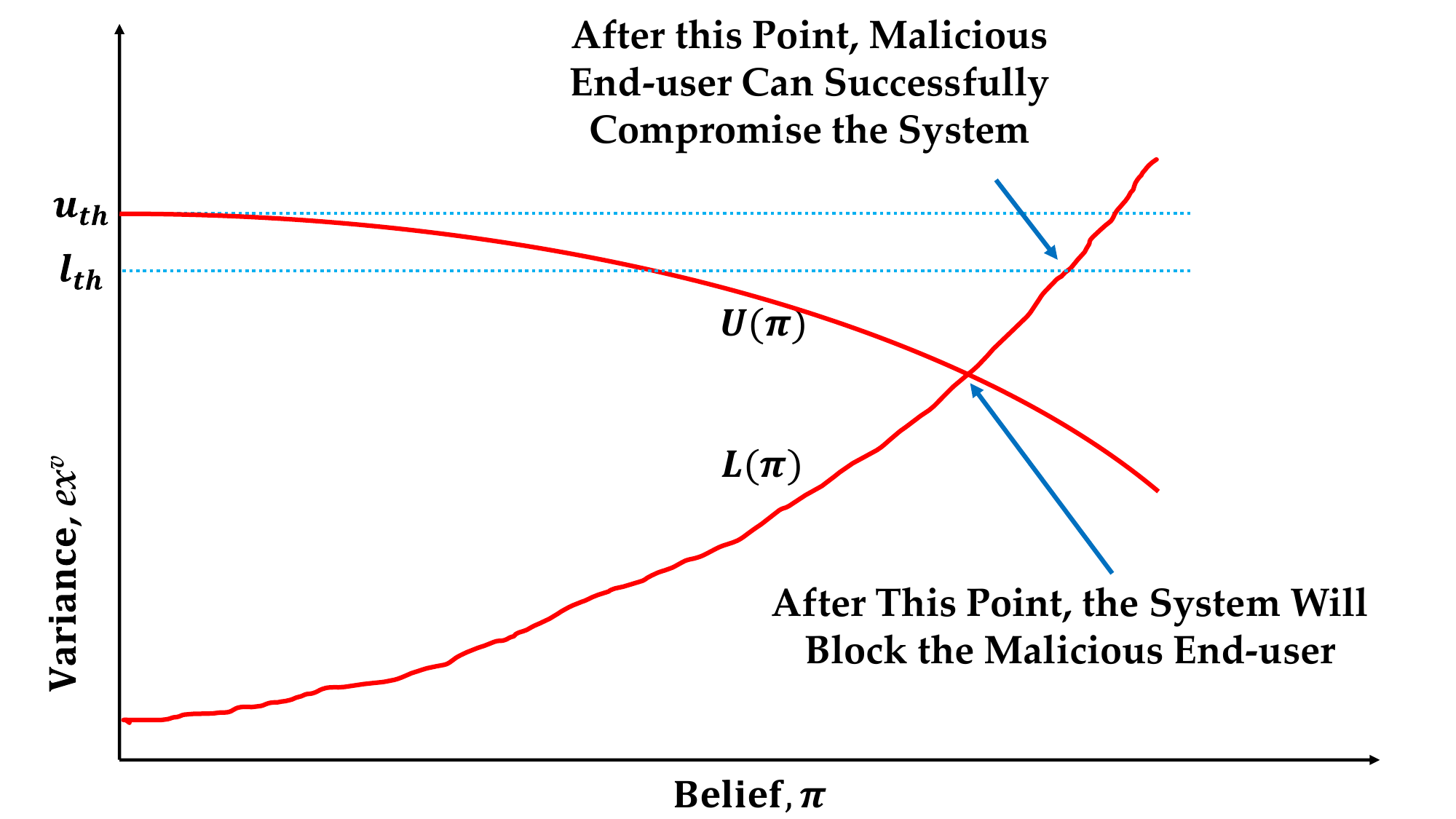}
\caption{Illustration of a continuous path analysis of $U(\pi)$ and $L(\pi)$ in Markov Perfect Equilibrium}
\label{uandl}\vspace{-0.3cm}
\end{figure}
%To characterize the MPE, we only consider the payoff relevant variables at any time, i.e., the current variance value $EX_{t}^{v}$ and the current belief held by the \texttt{system} about the \texttt{end-user}'s type $\pi_{t}$. 
%Our focus is on MPE, where $(ex,pi) \to R_{+} \times [0,1]$ serves as a state variable pair. We use $R_{+}$ to denote the strictly positive reals, $(0,\infty)$. In this paper, we use uppercase $EX$ for stochastic process and $ex$ for the realizations of these. 
%\mj{MJ-Start here..}
%We impose the following restrictions on belief-updating after off-path events: After any decision by the \texttt{end-user} to stop , the \texttt{system} believes that \texttt{end-user} is of type $m$ if the explanation generated for that \texttt{end-user} has reached the maximum relevant explanation value (defined later). This includes the case where the \texttt{end-user} has earned a reputation of $\pi_{t}= 1$, but is actually type $m$ and chooses to stop signaling. Moreover, if the \texttt{end-user} is ever believed to be malicious with certainty ($\pi_{t} = 0$), he can never recover; reputation of that \texttt{end-user} remains at $\pi_{t} = 0$. 
%[D1 criteria is an equilibrium refinement that requires out-of-equilibrium beliefs to be supported on types that have the most to gain from deviating from a fixed equilibrium.]
Given this equilibrium concept, our main result is the characterization of querying activity of the malicious \texttt{end-user} and detection (stopping) strategies by the \texttt{system} in a unique equilibrium.
We assume that a decision to stop querying (i.e., deviating from honest behavior) is the last action in the game taken by the (malicious) \texttt{end-user}. This decision allows the \texttt{end-user} to either achieve the target of compromising the \texttt{system} and then getting blocked by it or getting blocked without reaching this target at all. In either case, the \texttt{system}'s belief about this \texttt{end-user} will jump to $\pi_{t} = 0$. The \texttt{end-user} has no further action, and the game reduces to a straightforward \emph{stopping problem} for the \texttt{system} i.e., the \texttt{system} decides when to stop the game. In that case, the continuation payoffs from that point on can be interpreted as the termination payoffs of the original signaling game.

Next, consider the state of the game before the \texttt{end-user} deviates/reveals and before the \texttt{system}'s block action. Since malicious \texttt{end-user} plays a mixed strategy that occurs \emph{on-equilibrium path}, \texttt{system}'s belief about the \texttt{end-user} evolves over time. Thus, a unique MPE consists of a state variable process $(EX_{t}^{v}, \pi_{t})$ and two cutoff functions, a non-increasing variance function $U(\pi_{t})$ for the \texttt{system} and an increasing variance function $L(\pi_{t})$ for the \texttt{end-user}, where
\begin{itemize}[leftmargin=*]
    \item The \texttt{system} immediately blocks the \texttt{end-user} if $ex^{v}_{s_{y}, t} \geq U(\pi_{t})$, i.e., $\tau = inf\{t>=0 : ex^{v}_{s_{y}, t} \geq U(\pi_{t})\}$. 
    \item The malicious \texttt{end-user} keeps querying for explanation (thus, its variance), whenever $ex^{v}_{e_{u}, t} < L(\pi_{t})$ and mixes between querying and not querying whenever $ex^{v}_{e_{u}, t} \geq L(\pi_{t})$, so that the curve $\{(L(\pi),\pi): \pi \in [0,1]\}$ serves as a \emph{reflecting boundary} for the process $(ex^{v}_{e_{u}, t}, \pi_{t})$.
\end{itemize}

We call such a unique MPE equilibrium a \emph{$(U,L)$ equilibrium}.
The first condition defines an upper boundary which tells the \texttt{system} that if an explanation variance value $ex^{v}_{s_{y}, t}$ at time $t$ (corresponding to a query sent by the \texttt{end-user}) is greater than or equal to this boundary ($U(\pi_{t})$), then the \texttt{end-user} is trying to compromise the \texttt{system}. In this case, the \texttt{system} should block the \texttt{end-user}.
The second condition above guides the behavior of the malicious \texttt{end-user}. Function $L(\pi_{t})$ represents the upper-bound of the target explanation variance value the malicious \texttt{end-user} wants to achieve given a certain belief $\pi_{t}$ at time $t$. When the explanation variance value corresponding to a query by an \texttt{end-user} is less than this boundary function, i.e., $ex^{v}_{e_{u}, t} < L(\pi_{t})$, then it is strategically better for the malicious \texttt{end-user} to keep querying (i.e., looks honest from \texttt{system's} perspective). However, if $ex^{v}_{e_{u}, t} \geq L(\pi_{t})$ then the malicious \texttt{end-user} has an incentive to stop querying. For the malicious \texttt{end-user}, this condition also represents that it is near to the desired (or target) variance threshold value - one more step by the malicious \texttt{end-user} can either lead to success (compromise of the \texttt{system}) or failure (getting blocked by the \texttt{system} before achieving its goal). 

For an intuition on the structure of the MPE in our setting, suppose that the current belief is $\pi_{t}$. If the variance computed for an explanation generated by an explanation method is sufficiently large, i.e., close to a threshold variance value, then the \texttt{system} should block the \texttt{end-user} who is suspiciously trying to move in the direction of the model classification boundary. This cutoff for the variance is a non-increasing function of $\pi$ because, by definition, the \texttt{end-user} is less likely to be honest when the variance value is sufficiently close to the threshold value and $\pi_{t}$ is large. Thus, when threshold value becomes greater than or equal to the optimal function $U(\pi_{t})$ at any time $t$, \texttt{system} will block the \texttt{end-user}. This is intuitively shown in Figure \ref{uandl}, where $u_{th}$ represents the variance threshold for an explanation generated by the \texttt{system} and $l_{th}$ represents the variance threshold for the explanation after the \texttt{system} adds noise based on its belief, which can be given to the \texttt{end-user}. [0, $u_{th}] $ or [0, $l_{th}$] represents the pooling region where an MPE can occur, if \texttt{end-user} is not blocked by the \texttt{system}. 
%The reason being if $L(\pi_{t}) = l_{th}$ and $ex_{t}^{v} \geq L(\pi_{t})$ for the \texttt{end-user} or $U(\pi_{t}) = u_{th}$ and $ex_{t}^{v} \geq U(\pi_{t})$ for the \texttt{system}, deviation from the honest behavior will be the last-step a malicious \texttt{end-user} can take to launch the attack the \texttt{system}. Thus, an MPE will always occur in either range: [0, $u_{th}] $ or [0, $l_{th}$]. 

%Next, we argue that \texttt{end-user} types must separate above some upper cutoff of $L(\pi_{t})$. % In this paper, we have assumed that equilibrium occurs when both types of end user pool%. 
%Suppose that initially both types pool and in particular, type $m$ always signals (equilibrium condition). Then starting from the state $(ex_{t}, \pi_{t})$ = $(ex_{0}, \pi)$, as no information arrives about the \texttt{end-user}'s type, the belief $\pi_{t}$ holds constant at $\pi$. For sufficiently large $ex_{t}$, if the malicious \texttt{end-user} is very close to the desired target, then she/he would strictly prefer to deviate in order to compromise the \texttt{system} now, contradicting equilibrium. Thus, deviation must be in the support of type $m$'s strategy above some cutoff $L(\pi_{t})$.

\section{Equilibrium Analysis}
\label{sec:analysis}
Next, we try to analyze conditions under which a unique MPE exists in the game described above, i.e., we try to construct a $(U,L)$ equilibrium by finding conditions under which optimal functions $L(\pi_{t})$ and $U(\pi_{t})$ exist.
%, given the state of the game at any time $t$. 

\noindent
\textbf{High-level Idea:} As mentioned in section \ref{sec:model}, an MPE is defined as a pair of functions $L(\pi_{t})$ and $U(\pi_{t})$. Thus, we first need to show that these two optimal functions exists. To prove that $L(\pi_{t})$ and $U(\pi_{t})$ exist, we prove the continuity and differentiability properties of $L(\pi_{t})$ (theorem \ref{thm:optU}) and  $U(\pi_{t})$ (theorem \ref{thm:optL}) in the belief domain ($\pi_{t} \in [0,1 ]$). As we have assumed that the \texttt{system} plays a pure strategy, we consider that $U(\pi_{t})$ (for the \texttt{system}) is optimal, and show that it exists and is continuous. 
%\mj{this sentence is not clear - need to discuss. What is optimal time strategy?}. 
%However, for the \texttt{end-user} we have assumed that he/she plays a randomized strategy, and thus, is unsure of the optimal $L(\pi_{t})$. 
However, computing an optimal $L(\pi_{t})$ is non-trivial, as we have assumed that the \texttt{end-user} plays a randomized strategy.
%\mj{check this sentence for correctness}.
% using ordinary differential calculus, 
Therefore, to compute $L(\pi_{t})$, we first construct two bounding functions $L^{+}(\pi_{t})$ (upper) and $L^{-}(\pi_{t})$ (lower) and show 
%again use ordinary differential calculus to show that these two 
that such functions exists (in lemmas \ref{lemma:lplus} and \ref{lemma:lminus}, respectively). To compute these functions we use the boundary conditions (section \ref{subsec:ctbw}) at the decision boundaries, i.e., Pooling, Separating, and Detection and Block (outlined in Table \ref{tab:payoffs}) and the value functions defined below.
%\mj{check}.
%For the MPE characterization and to show that a unique MPE exists in the game, 
We then show that as $\pi_{t}$ increases, $L^{+}(\pi_{t})$ and and $L^{-}(\pi_{t})$ will converge in the range $(0, u_{th})$ (lemma \ref{lemm:uth}) or $(0, l_{th})$ (lemma \ref{lemm:lth}) as we have considered that an MPE occurs when both types of \texttt{end-user} pool.  Finally, we show that the first intersection point (root) of $L^{+}(\pi_{t})$ and $L^{-}(\pi_{t})$ is a unique MPE, if the \texttt{end-user} is not blocked by the \texttt{system} before that point.  
%(as outlined in Section \ref{subsec:setup} payoffs) . 
At these decision boundaries, both players decide to either continue or stop.
%Therefore, the optimal choice for both the \texttt{system} and the \texttt{end-user} is a binary variable: 0 to stop and 1 to continue \mj{is this last sentence needed? can be combined with previous sentence?}. 
%and we have written down value matching, smooth pasting and reversed explanation condition, i.e., for example, if \texttt{end-user} was expecting $L^{+}(\pi_{t})$, but he/she received $L^{-}(\pi_{t})$, at the possible decision boundaries and decided if a player wants to continue or stop for optimality. 

\subsection{Value Functions}
\label{subsec:valuefunctions}
As mentioned earlier, we are considering an equilibrium that occurs when both types of \texttt{end-user} pool, i.e., $ex^{v}_{e_{u}, t}$ is strictly below $L(\pi_{t})$. In this pooling scenario, no information becomes available to the \texttt{system} about the type of the \texttt{end-user}; thus, belief ($\pi_{t}$) remains constant. Hence, we write the value functions for both the players conditioned on no deviation by the \texttt{end-user}. 
%As mentioned in section 2, the value function ($H$) of any player is given by:
%
%$$rH(s, t) = f(s,u^{*},t) + \frac{\partial H}{\partial t} + \frac{\partial H}{\partial s}a(s,u^{*},t) + \frac{1}{2}\frac{\partial^{2} H}{\partial s^{2}}b(s,u^{*},t)^{2}$$
As we consider an infinite horizon in our game, there is no known terminal (final) value function. Hence, these value functions are independent w.r.t. $t \in R$, as $t$ singularly has no effect on them. 
The \texttt{end-user}'s value function ($F$) should solve the following HJB equation representing his/her risk-less return:\vspace{-0.1cm}
{
$$rF(ex^{v}_{e_{u}},\pi) = \mu ex^{v}_{e_{u}} F^{'}_{ex^{v}_{e_{u}}}(ex^{v}_{e_{u}}, \pi) $$\vspace{-0.4cm}
$$+ \frac{1}{2}\sigma^{2}  (ex^{v}_{e_{u}})^{2}  F^{''}_{ex^{v}_{e_{u}}}(ex^{v}_{e_{u}}, \pi) + \psi ex^{v}_{e_{u}}$$

}
where $F^{'}_{ex^{v}_{e_{u}}}$ and $F^{''}_{ex^{v}_{e_{u}}}$ are the first and second order partial derivative of the value function $F(ex^{v}_{e_{u}},\pi)$ w.r.t. $ex^{v}_{e_{u}}$, respectively, and,
%, where $(ex,\pi)$ is the state of our game. 
$\mu$ and $\sigma$ are the drift/mean and the variance/volatility of the variance process $EX_{t}^{v}$, respectively.  
$\psi$ is the payoff coefficient which depends on the stage payoffs of the \texttt{end-user},
%- signaling, not signaling and detection/block stage 
as represented in Table \ref{tab:payoffs}.
% Thus, $f(s,u(s,t),t) = f(ex,\pi) = \psi ex$.
%MJ - not sure if the below line is required! 
%The $rF(ex,\pi)$ in the above equation represents the risk-less return of the \texttt{end-user} and the right-hand side is the expected utility/return plus the flow payoff given the state of the game. 
The solution to the above equation can be represented as:
$$ F(ex^{v}_{e_{u}}, \pi) = A_{1}(\pi) (ex^{v}_{e_{u}})^{\beta_{1}} + A_{2}(\pi) (ex^{v}_{e_{u}})^{\beta_{2}} + \frac{\psi ex^{v}_{e_{u}}}{r-\mu}$$
for some constants $A_{1}(\pi)$ and $A_{2}(\pi)$, where $\beta_{1} > 1$ and $\beta_{2} < 0$ are the roots of the characteristic equation \cite{dixit2012investment}. 
Similarly, the \texttt{system}'s value function $V(ex^{v}_{s_{y}},\pi)$ should satisfy the following equation, conditioned on  $ex^{v}_{e_{u}} < L(\pi)$ and $\pi$ staying constant:\vspace{-0.1cm}
$$rV(ex^{v}_{s_{y}},\pi) = \mu ex^{v}_{s_{y}} V^{'}_{ex^{v}_{s_{y}}}(ex^{v}_{s_{y}}, \pi) $$ \vspace{-0.4cm}
$$+ \frac{1}{2}\sigma^{2}  (ex^{v}_{s_{y}})^{2} V^{''}_{ex^{v}_{s_{y}}}(ex^{v}_{s_{y}}, \pi) + \psi ex^{v}_{s_{y}}$$
where $V^{'}_{ex^{v}_{s_{y}}}$ and $V^{''}_{ex^{v}_{s_{y}}}$ are the first and second order partial derivative of the value function $V(ex^{v}_{s_{y}},\pi)$ w.r.t. $ex^{v}_{s_{y}}$, respectively. As before, $\psi$ is the payoff coefficient which depends on the stage payoffs of the \texttt{system} as shown in Table \ref{tab:payoffs}. 
%The characterization of the above equation is the same as that of \texttt{end-user's} Hamilton-Jacobi-Bellman equation. 
The solution to the above equation can be represented as:
$$ V(ex^{v}_{s_{y}}, \pi) = B_{1}(\pi) \times (ex^{v}_{s_{y}})^{\beta_{1}} + B_{2}(\pi)  (ex^{v}_{s_{y}})^{\beta_{2}} + \frac{\psi ex^{v}_{s_{y}}}{r-\mu}$$
for some constant $B_{1}(\pi)$ and $B_{2}(\pi)$. We will use different boundary conditions to determine $A_{1}(\pi)$, $A_{2}(\pi)$, $B_{1}(\pi)$ and $B_{2}(\pi)$. Then, we will use these conditions to determine $U(\pi_{t})$ and $L(\pi_{t})$.

\subsection{Analytical Results} 
\label{subsec:ana;ytical}
%Our main analysis is based on finding conditions under which a unique Markov Perfect Equilibrium occurs in the game.
%As mentioned in \ref{subsec:eqdesc}, we have assumed that the equilibrium occurs only when both types of \texttt{end-user}'s pool. Moreover, malicious \texttt{end-user} will pool (or keep querying the \texttt{system}) with the honest \texttt{end-user} with high probability (there is a chance, \texttt{end-user} may deviate earlier because he/she is playing a mixed strategy) only if the explanation variance threshold is not reached. 

Our first aim is to compute the \texttt{system}'s threshold $u_{th}$ (Lemma \ref{lemm:uth}) and the corresponding \texttt{end-user}'s threshold $l_{th}$ (Lemma \ref{lemm:lth}). These thresholds define the region in which an MPE may occur in the game, if the necessary conditions are satisfied. Therefore, these two lemmas will be used to determine if a unique MPE exists in the game or not. 
Due to space constraints, we have moved all proofs to the supplementary document.

%The first two results (Lemma \ref{lemm:uth} and \ref{lemm:lth}) determines the \texttt{system}'s thresholds $u_{th}$ and the \texttt{end-user}'s threshold $l_{th}$, respectively, as described earlier in Section \ref{subsec:eqdesc}. 
%$u_{th}$  is the  maximum relevance explanation which can be given by the \texttt{system} to the \texttt{end-user} and $l_{th}$  is the  minimum relevance explanation which can be given by the \texttt{system} to the \texttt{end-user}.
%These two lemmas are used later to define the valid range in which a unique root of the polynomial exists \mj{this last sentence does not make much sense by itself}. 

\begin{lemma} There exists a positive upper bound $u_{th}$ on the variance of an explanation generated by an explanation method representing the maximum variance value that can be reached for the query sent by the \texttt{end-user}. 
\label{lemm:uth}
\end{lemma}
%\begin{proof}
%As stated in Appendix \ref{app:lemma1}.
%\end{proof}
    
\begin{lemma} There exists a positive upper bound $l_{th}$ on the variance of an explanation given by the \texttt{system} to the \texttt{end-user} representing maximum variance value needed to be reached by the \texttt{end-user} to compromise the \texttt{system}.
\label{lemm:lth}
\end{lemma}
%\input{lemma2}
%\begin{proof}
%As stated in Appendix \ref{app:lemma2}.
%\end{proof}

Our next aim is to characterize the two optimal cutoff functions, i.e., \texttt{system}'s function $U(\pi_{t})$ and \texttt{end-user}'s function $L(\pi_{t})$, as these two functions represent the MPE in the game. These cut-off functions help the \texttt{system} and the \texttt{end-user} to play optimally in each stage of the game. For example, if the \texttt{system} doesn't have any knowledge of $U(\pi_{t})$, then it won't know the range of the variance values being computed for the explanations, which are given to the \texttt{end-user} after adding some noise based on its belief. Hence, an adversary will be able to easily compromise the \texttt{system}. In contrast, $L(\pi_{t})$ function knowledge will guide an adversary on how to optimally compromise the \texttt{system}. For that reason, we first prove that $U(\pi_{t})$ exists and is non-increasing and continuously differentiable (Theorem \ref{thm:optU}\footnote{In order to simplify the exposition of the proofs, we have replaced $L(\pi_{t})$ with $L(\pi)$ and $U(\pi_{t})$ with $U(\pi)$ in these results.}).
%The next set of results show that an optimal explanation path exists for both the \texttt{system} ($U(\pi_{t})$) and the \texttt{end-user} ($L(\pi_{t})$) by proving $U(\pi_{t})$ is non-increasing and continuously differentiable (Theorem \ref{thm:optU}) and by proving $L(\pi_{t})$  increasing and continuously differentiable (Theorem \ref{thm:optL}). 
%For the \texttt{system}, we will simply prove that there exists an optimal function $U(\pi_{t})$ which is non-increasing and continuously differentiable. 
%For the \texttt{end-user}, we also prove that there exists two cut-off functions $L^{+}(\pi_{t})$ (explanation with maximum relevance given by the \texttt{system}) and $L^{-}(\pi_{t})$ (explanation with minimum relevance given by the \texttt{system}), which are both increasing and continuously differentiable (Lemmas \ref{lemma:lplus} and \ref{lemma:lminus}). Finally, we show (Theorem \ref{thm:converge}) that there exists a unique root of the polynomial \mj{which polynomial} representing the convergence of $L^{+}(\pi_{t})$ and $L^{-}(\pi_{t})$ at some continuous time $t$, and this intersection point represents.....\mj{complete}.
\begin{theorem}
\label{thm:optU}
$U(\pi_{t})$ is non-increasing and continuously differentiable function in domain $[0, 1]$ if and only if either $\beta_{2}\beta_{1} J^{`}(\pi, t)^{\beta_{2}-1} \leq \beta_{1}\beta_{2} J^{`}(\pi, t)^{\beta_{1}-1}$ or $\beta_{2}(\beta_{1}-1) J^{`}(\pi, t)^{\beta_{2}-1} \leq \beta_{1}(\beta_{2}-1) J^{`}(\pi, t)^{\beta_{1}-1}$, where $J(\pi, t) = \frac{L(\pi)}{U(\pi)}$.
\end{theorem}
%\begin{proof}
%As stated in Appendix \ref{app:theorem1}.
%\end{proof}
%\input{theorem1}

To prove $L(\pi)$ (Theorem \ref{thm:optL}) exists and is increasing and continuously differentiable, we first characterize an explanation variance path $L^{+}(\pi)$ (Lemmas \ref{lemma:lplus}), which represents the maximum variance values that can be computed by the \texttt{end-user} for the given explanations, and a variance path $L^{-}(\pi)$ (Lemma \ref{lemma:lminus}), which represents the minimum variance values for the explanations given by the \texttt{system} to the \texttt{end-user}. We write three equations each for $L^{+}(\pi)$ and $L^{-}(z)$ according to the value matching, smooth pasting, and the condition in which the variance of the explanation received is opposite of what \texttt{end-user} expected. Then, we demonstrate that both these functions are increasing and continuously differentiable. The purpose for doing this is to use these lemmas to show that as $\pi \rightarrow 1$, both $L^{+}(\pi)$ and $L^{-}(\pi)$ starts to converge and becomes equal to $L(\pi)$ after some point.

\begin{lemma}
\label{lemma:lplus}
   $L^{+}(\pi)$ is a well-defined, increasing, continuous and differentiable function in domain $[0, 1]$ if and only if  $\lambda^{'}(L^{+}(\pi), \pi) > 0$ and $P>0$, where $\lambda()$ is the termination payoff if the \texttt{end-user} decides to deviate and attack the \texttt{system}.
\end{lemma}
%As stated in Appendix \ref{app:lemma3}.
%\input{lemma3}
%    
%\begin{proof}
%As stated in Appendix \ref{app:lemma3}.
%\end{proof}

\begin{lemma}
\label{lemma:lminus}
$L^{-}(\pi))$ is a well-defined, increasing, continuous and differentiable function in domain $[0, 1]$ if and only if either $(\frac{\partial A_{1}^{+}(z)}{\partial \pi} L^{-}(\pi)^{\beta_{1}} + \frac{\partial A_{2}^{+}(\pi)}{\partial \pi} L^{-}(\pi)^{\beta_{2}})< 0$ or $(A_{1}^{+}(\pi) \beta_{1} L^{-}(\pi)^{\beta_{1}-1} + A_{2}^{+}(\pi) \beta_{2} L^{-}(\pi)^{\beta_{2}-1})< 0 $.
\end{lemma}
%As stated in Appendix \ref{app:lemma4}.
%\input{lemma4}
%  
%\begin{proof}
%As stated in Appendix \ref{app:lemma4}.
%\end{proof}

\begin{theorem}
\label{thm:optL}
       $L(\pi)$ is a well-defined, increasing, continuous and differentiable function domain $[0, 1]$ if and only if either $\lambda^{'}(L(\pi), \pi) > 0$ and $P>0$.
\end{theorem}
%As stated in Appendix \ref{app:theorem2}.
%\input{theorem2}
%  
%\begin{proof}
%As stated in Appendix \ref{app:theorem2}.
%\end{proof}

Finally, we show that such a point where $L^{+}(\pi_{t})$ and $L^{-}(\pi_{t})$ converge (or intersect) exists, and thus, a unique MPE (Theorem \ref{thm:converge}) exists in the game. 
\begin{theorem}
\label{thm:converge}
A unique MPE or a point, $ \varsigma = \frac{\lambda(L^{+}(\pi), \pi) \times (r-\mu)}{P \times L^{-}(\pi)}$, exists in the game where the two curves $L^{+}(\pi)$ and $L^{-}(\pi)$ starts to converge, if and only if $\frac{\beta_{2}}{\varsigma^{\beta_{2} + 1}}\left[L^{+}(\lambda^{'} - \frac{P}{r-\mu}) - \beta_{1} (\lambda - \frac{P L^{+}}{r-\mu})\right] \geq  \frac{\beta_{1}}{\varsigma^{\beta_{1} + 1}}\left[ \beta_{2}(\lambda - \frac{P L^{+}}{r-\mu}) - L^{+} (\lambda^{'} -\frac{P}{r-\mu})\right]$.
\end{theorem}
%As stated in Appendix \ref{app:theorem3}.
%\input{theorem3}
%\begin{proof}
%As stated in Appendix \ref{app:theorem3}.
%\end{proof}

\section{Experimental Setup}
\label{sec:setup}
%\vspace{-0.2cm}

We employ the \emph{Captum} \cite{kokhlikyan2020captum} framework to generate four different types of explanations: \emph{Gradient*Input}, \emph{Integrated Gradients}, \emph{LRP}, and \emph{Guided Backpropagation}. We use \emph{PyTorch} framework to conduct the training and attack-related experiments. \emph{Gradient*Input} is used as the baseline (to compare the results of the other explanation methods). 
%As mentioned in \cite{shokri2021privacy}, high explanation variance values at any datapoint $\overrightarrow{x}$ indicates that $\overrightarrow{x}$ was not used for training the model, %the model is uncertain about the prediction of $\overrightarrow{x}$ paving the way towards a potential attack by an adversary.thus, 
Our objective in all the experiments is to analyze and compute the impact of different settings (factors) that can impact the capability of an adversary to launch MIA. We assume that when the game ends, both the \texttt{system} and the \texttt{end-user} will have access to their optimal strategies, thus the optimal explanation variance threshold ($u_{th}$ for the \texttt{system} and $l_{th}$ for the \texttt{end-user}). Hence, an adversary can use its optimal strategy and optimal threshold to conduct MIA, or a \texttt{system} can use its optimal strategy and optimal threshold to protect against MIA. As a result, we focus on 
two evaluation objectives in our experiments: (i) \emph{game evolution}, and (ii) \emph{MIA accuracy}. For the game evolution, we simulate and generate the future explanation variances for $t=100$ stages, according to the expression:
%\vspace{-0.2cm}
\begin{equation}
    EX_{t}^{v} = EX_{0}^{v} * e ^{\left((\mu-\frac{1}{2}\sigma^{2}) + \sigma W_{t}\right)} \label{eqn:gbm}
    \vspace{-0.3cm}
\end{equation}

%We will first use the \emph{Gradient*Input} to generate explanations and examine the existence of MPE in each of the dataset outlined below.
The above equation is the solution to the GBM (Equation \ref{eqn:gbm}) of $EX_{t}^{v}$, derived using the \textit{it$\hat o$'s} calculus \cite{dixit2012investment}. 
%$\mu$ is the constant \emph{drift} and $\sigma > 0$ is the constant \emph{volatility} of the variance process $EX_{t}^{v}$, and
$\mu$ and $\sigma > 0$ are computed using the variance generated for the test datapoints for each of the dataset. 
%$EX_{0}^{v}$ is the initial explanation variance computed at $t=0$.
In our experiments, we take $EX_{0}^{v}$ as the last index value of the test data points' generated explanation variance, as we use this initial value to generate future explanations. 
Using the obtained optimal strategies and thresholds, we compute the attack accuracy in terms of the attacker's success rate in launching the MIA or the accuracy of the \texttt{system} in preventing the MIA. 
%\mj{check-I edited this sentence}.
%After the analysis of the game or when the game ends, both the \texttt{system} and the \texttt{end-user} will have access to their optimal strategies, thus the optimal explanation variance threshold ($u_{th}$ for the \texttt{system} and $l_{th}$ for the \texttt{end-user}). 

%Below we provide the configurations of the different datasets used in the game evaluation and the metric we employ to measure the accuracy of the attacks. 

\noindent
\textbf{Datasets. }We use five popular benchmark datasets on which we perform our game analysis and attack accuracy evaluations: Purchase and Texas datasets \cite{nasr2018machine}, CIFAR-10 and CIFAR-100 \cite{sablayrolles2019white}, and the Adult Census dataset \cite{Dua:2019}. To ease the comparison, the setup and Neural Network (NN) architectures are aligned with existing work on explanation-based threshold attacks \cite{shokri2021privacy}. Table \ref{tab:datasets} details each dataset's configuration. Below, we briefly provide details about each of the datasets. 

\emph{Purchase:} This dataset is downloaded from Kaggle's ``Acquire Valued Shoppers Challenge" \cite{shoppers}. The challenge aims to predict customers that will make a repeat purchase. The dataset has $197,324$ customer records, with each record having $600$ binary features. The prediction task is to assign each customer to one of the $100$ labels. We use the same NN architecture used by Shokri et al. \cite{shokri2021privacy}, a four-layer fully connected neural network with \textit{tanh} activations and layer sizes equal to $1024$, $512$, $256$ and $100$. We use \textit{Adagrad} optimizer with a learning rate of $0.01$ and a learning rate decay of $10^{-7}$ to train the model for $50$ epochs.

\emph{Texas:} This dataset contains patient discharge records from Texas hospitals \cite{texas}, spanning years 2006 through 2009. It has been used to predict the primary procedures of a patient based on other attributes such as hospital id, gender, length of the stay, diagnosis, etc. This dataset is filtered to incorporate only 100 classes (procedures) and has 67,330 records. The features are transformed to include only binary values, such that each record has 6,170 features. 
We use a five-layer fully connected NN architecture with \textit{tanh} activations and layer sizes of 2048, 1024, 512, 256 and 100. We employ the \textit{Adagrad} optimizer with a learning rate of 0.01 and a learning rate decay of $10^{-7}$ to train the model for 50 epochs.

\emph{CIFAR-10 and CIFAR-100:} CIFAR-10 and CIFAR-100 are the benchmark datasets for image classification tasks. Each dataset consists of 50K training and 10K test images.
As NN architecture, we designed two convolutional layers with max-pooling and two dense layers, all with \textit{tanh} activations for CIFAR-10 and using Alexnet \cite{krizhevsky2012imagenet} for CIFAR-100. We train the model for 50 epochs with learning rates of 0.001 and 0.0001 for CIFAR-10 and CIFAR-100, respectively. In both cases, we use \textit{Adam} optimizer to update the parameters of the model.

\emph{Adult:} This dataset is downloaded from the 1994 US Census database \cite{Dua:2019}. It has been used to predict whether a person's yearly income is above 50,000 USD or not. It has 48,842 records and a total of 14 features, including categorical and numeric features. After transforming categorical features into binary features, each data point resulted in 104 features. A dataset of 5,000 data points is sampled from the original dataset because of the large size, containing 2,500 training data points and 2,500 testing data points. 
We use a five-layer fully-connected NN architecture with \textit{tanh} activations and layer sizes equal to 20, 20, 20, 20 and 1. We use an \textit{Adagrad} optimizer with a learning rate of 0.01 and a learning rate decay of $10^{-7}$ to train the model for 50 epochs.

\begin{table}[t]
\centering
%\caption{Configuration of the datasets utilized in the analysis of the game.}
\caption{Dataset Configurations.}
\vspace{-0.2cm}
\resizebox{7cm}{!}{
\begin{tabular}{| c | c |c | c | c |} 
 \hline
\textbf{Datasets} & \textbf{Points} & $\#$\textbf{Features} & \textbf{Type} & $\#$\textbf{Classes} \\  \hline
 Purchase             & 197,324           & 600 					  & Binary            & 100  \\ \hline
 Texas 			  & 67,330 		  & 6,170 				  & Binary            & 100 \\ \hline
 CIFAR-100   	  & 60,000    	  & 3,072				  & Image 		 & 100 \\ \hline
 CIFAR-10      	  & 60,000       	  & 3,072 				  & Image            & 2 \\ \hline
  Adult			  & 48,842		  & 24					  & Mixed 		 & 2  \\ \hline
\end{tabular}
}
\label{tab:datasets}\vspace{-0.2cm}
%\vspace{-0.4cm}
\end{table}

\textbf{Evaluation metric. }We compute \textit{True Positive Rate (TPR)} metric to estimate MIA accuracy after the game has ended and each player has formulated the best response strategy concerning the other player. TPR specifies how accurately an attacker can infer the membership of the data points. \changed{We consider training data points to test against the optimal strategy of the \texttt{system}. Since the sample space that we have considered contains only actual training members, there can be only two outcomes: correctly classified and incorrectly classified.}
%After the game analysis, as mentioned earlier, we assume that both the \texttt{system} and the \texttt{end-user} will have access to their optimal strategies, i.e., will have access to the explanation variance upper thresholds $u_{th}$ and $l_{th}$, respectively. 
%first, we compute the explanation variance of the training set, and then, we compare the obtained explanation variance with the \texttt{system}'s $u_{th}$ to determine how many points explanation variance lies below $u_{th}$ as mentioned in Section \ref{subsec:mia}.
%\mj{This is NOT CLEAR AT ALL. Let's discuss this!}. 
The total number of training data points correctly inferred as training points (using $u_{th}$) are called True Positives ($TP$), while the number of training members discerned as non-training members are called False Negatives ($FN$). Thus, $TPR = \frac{TP}{TP+FN}$. 

\section{Evaluation}
\label{sec:eval}

In this section, we present experimental analyses of our game model, wherein we compute the impact of different settings on two objectives: (i) the game or equilibrium evolution to obtain optimal strategies and (ii) attack accuracy, i.e., TPR. 
We begin by analyzing the impact of the \emph{Gradient*Input} explanations (baseline) on the game equilibrium and attack accuracy for all the considered datasets, 
%mentioned in Section \ref{sec:setup} 
and then compare it with the results obtained for the other explanation methods. After that, we also analyze the influence of other factors (e.g., input dimension and overfitting) on the attack accuracy. 

\subsection{Impact of Different Attack Information Sources}
%In this section, we first analyze the game's evolution for the different datasets mentioned in the previous section for the baseline setting. Then, we compute the attack accuracy of the launched explanation-based threshold attacks (Section \ref{subsec:mia}) by determining the $TPR$. Lastly, we perform the same analysis for other explanation methods and compare the obtained results with the results of the baseline setting.
%In order to simulate the variance of the explanations as GBM, we first compute a list of the explanation variances as mentioned in Section \ref{sec:setup} for the test data points for each dataset. Then, we estimate the mean and standard deviation of explanations variance generated for the output labels and use that to 
As mentioned in Section \ref{sec:setup}, for each dataset, we first sample a series of (future) explanations by employing GBM (using Equation \ref{eqn:gbm}). 
%Noise to be added to the explanations variance provided by the \texttt{system} to the \texttt{end-user} is sampled from a Normal distribution with the same mean and standard deviation of the generated explanations variances. 
The sampled noise is added to the generated explanations variance based on the computed belief at that time (using Bayes' rule in Section \ref{subsec:setup}), such that larger the belief that the \texttt{end-user} is honest, the smaller the noise value added to the explanation, and vice-versa. 
%Game model parameters that we chose in our numerical simulations for all explanation techniques are outlined in Table \ref{fig:sys_parameters} in the Appendix.
Then, we compute different functional paths for the \texttt{system} and the \texttt{end-user} defined in Section \ref{subsec:eqdesc} using the closed form representation detailed in Section \ref{subsec:ana;ytical}, i.e., we compute $U(\pi_{t})$, $L^{+}(\pi_{t})$,  $L^{-}(\pi_{t})$ and  $L(\pi_{t})$ functions. 
% assumed in figure \ref{fig:sys_parameters} (in Appendix \ref{subsec:gp}) for each of the explanation method.  
The termination payoff, $\lambda(ex^{v}_{e_{u}},\pi_{t})$ (defined in \ref{subsec:ctbw}), which is used to write the boundary conditions in the computation of $L^{+}(\pi_{t})$,  $L^{-}(\pi_{t})$ and  $L(\pi_{t})$ (Lemma \ref{lemma:lplus} and Lemma \ref{lemma:lminus}, and Theorem \ref{thm:optL}) is assumed to be:
\begin{equation}
\lambda(ex^{v}_{e_{u}},\pi_{t}) = \frac{0.8 \times ex^{v}_{e_{u}} \times log({\pi_{t} \times 2}) + \pi_{t} \times ex^{v}_{e_{u}}}{b} \nonumber
\end{equation}
where, $ex^{v}_{e_{u}}$ is the value of any \texttt{end-user}'s functional path (considered for the specific computation) at time $t$, and $b$ is the model parameter set differently for each explanation method. The parameters to compute $\lambda(ex^{v}_{e_{u}},\pi_{t})$ are empirically chosen based on their suitability to each of the four explanation methods considered in the paper. Based on our numerical simulations, below are some important observations we make for each dataset, both in the baseline setting (i.e., for the \emph{Gradient*Input} method) and the other three gradient-based explanation techniques.

\begin{figure}[!ht]
     \centering
     \iffalse
      \begin{subfigure}[b]{0.23\textwidth}
         \centering
         \includegraphics[width=\textwidth]{Figures/cifar-10/ex_toeu_g_vs_tem_g_copy.pdf}
         \caption{CIFAR-10}
         \label{fig:cifar10_exge_vs_exgi}
     \end{subfigure}
     \fi
     \begin{subfigure}[b]{0.23\textwidth}
         \centering
         \includegraphics[width=\textwidth]{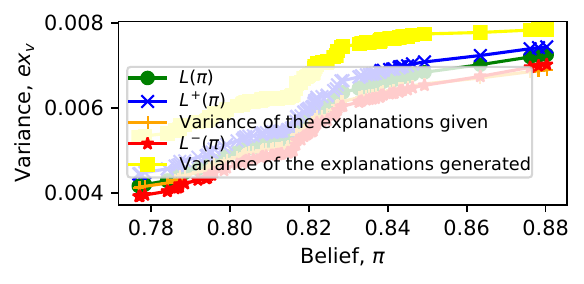}
         \caption{CIFAR-10}
         \label{fig:cifar10_diff_var_eu}
     \end{subfigure}
     \begin{subfigure}[b]{0.23\textwidth}
         \centering
         \includegraphics[width=\textwidth]{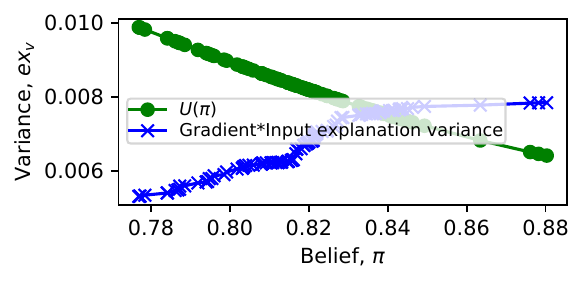}
         \caption{CIFAR-10}
         \label{fig:cifar10_diff_var_system}
     \end{subfigure}
     \iffalse
     \centering
      \begin{subfigure}[b]{0.23\textwidth}
         \centering
         \includegraphics[width=\textwidth]{Figures/cifar-100/ex_toeu_g_vs_tem_g_copy.pdf}
         \caption{CIFAR-100}
         \label{fig:cifar100_exge_vs_exgi}
     \end{subfigure}
     \fi
     \begin{subfigure}[b]{0.23\textwidth}
         \centering
         \includegraphics[width=\textwidth]{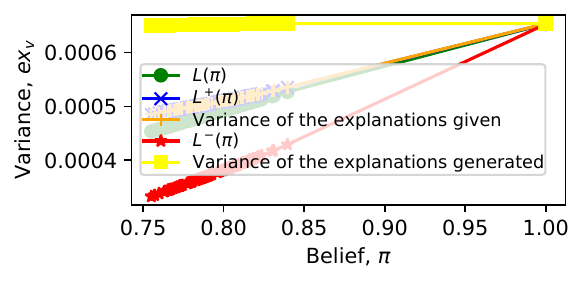}
         \caption{CIFAR-100}
         \label{fig:cifar100_diff_var_eu}
     \end{subfigure}
     \begin{subfigure}[b]{0.23\textwidth}
         \centering
         \includegraphics[width=\textwidth]{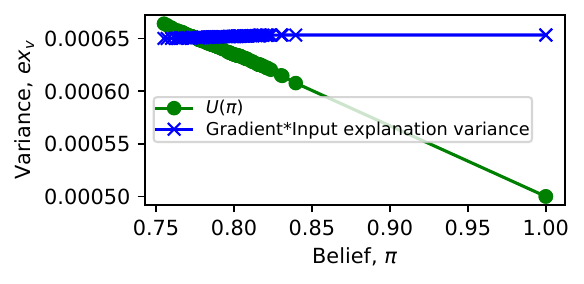}
         \caption{CIFAR-100}
         \label{fig:cifar100_diff_var_system}
     \end{subfigure}
    \iffalse
     %\centering
      \begin{subfigure}[b]{0.23\textwidth}
         \centering
         \includegraphics[width=\textwidth]{Figures/Adult/ex_toeu_g_vs_tem_g_copy.pdf}
         \caption{Adult}
         \label{fig:adult_exge_vs_exgi}
     \end{subfigure}
     \fi
     \begin{subfigure}[b]{0.23\textwidth}
         \centering
         \includegraphics[width=\textwidth]{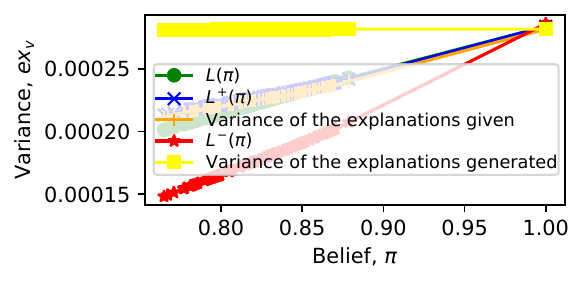}
         \caption{Adult}
         \label{fig:adult_diff_var_eu}
     \end{subfigure}
     \begin{subfigure}[b]{0.23\textwidth}
         \centering
         \includegraphics[width=\textwidth]{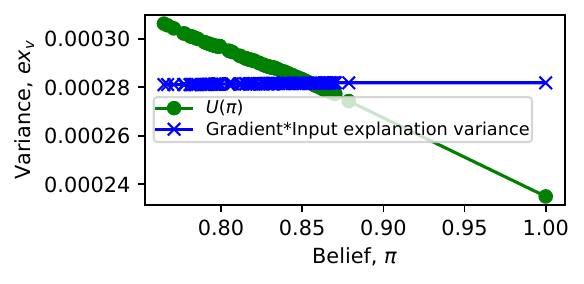}
         \caption{Adult}
         \label{fig:adult_diff_var_system}
     \end{subfigure}
     \iffalse
      \begin{subfigure}[b]{0.23\textwidth}
         \centering
         \includegraphics[width=\textwidth]{Figures/Purchase/ex_toeu_g_vs_tem_g_copy.pdf}
         \caption{Purchase}
         \label{fig:purchase_exge_vs_exgi}
     \end{subfigure}
     \fi
     \begin{subfigure}[b]{0.23\textwidth}
         \centering
         \includegraphics[width=\textwidth]{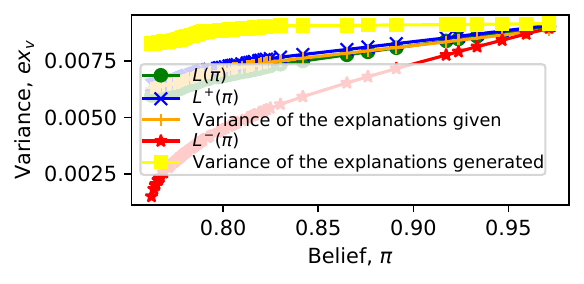}
         \caption{Purchase}
         \label{fig:purchase_diff_var_eu}
     \end{subfigure}
     \begin{subfigure}[b]{0.23\textwidth}
         \centering
         \includegraphics[width=\textwidth]{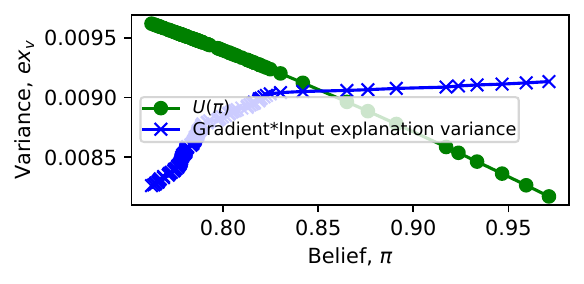}
         \caption{Purchase}
         \label{fig:purchase_diff_var_system}
     \end{subfigure}
     \iffalse
      \begin{subfigure}[b]{0.23\textwidth}
         \centering
         \includegraphics[width=\textwidth]{Figures/Texas/ex_toeu_g_vs_tem_g_copy.pdf}
         \caption{Texas}
         \label{fig:texas_exge_vs_exgi}
     \end{subfigure}
     \fi
     \begin{subfigure}[b]{0.23\textwidth}
         \centering
         \includegraphics[width=\textwidth]{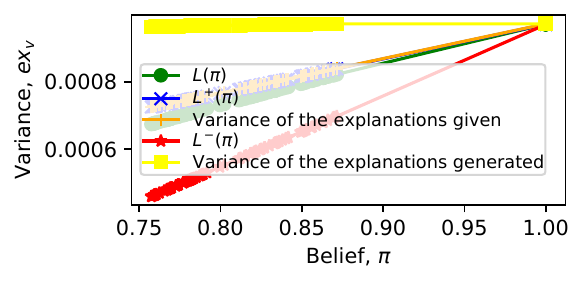}
         \caption{Texas}
         \label{fig:texas_diff_var_eu}
     \end{subfigure}
     \begin{subfigure}[b]{0.23\textwidth}
         \centering
         \includegraphics[width=\textwidth]{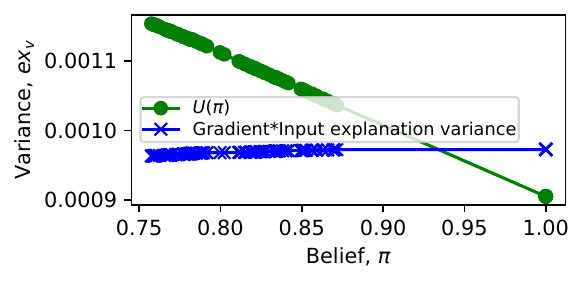}
         \caption{Texas}
         \label{fig:texas_diff_var_system}
     \end{subfigure}
        \caption{Different functional paths for the different datasets. (a), (c), (e), (g), and (i) represents the optimal functional paths for the \texttt{end-user}. (b), (d), (f), (h), and (j) represents the optimal functional paths for the \texttt{system}.}
        \label{fig:baseline_game}\vspace{-0.3cm} 
\end{figure}

\noindent
\textbf{Game Evolution in the Baseline Setting: }
Varying game evolution is realized for different datasets, as shown in Figure \ref{fig:baseline_game}. Below we analyze in detail the optimal paths obtained for each of the dataset. 
\begin{itemize}[leftmargin=*]
\item From the plots of the optimal functional path $U(\pi_{t})$ of the \texttt{system} for each of the dataset, as shown in Figures \ref{fig:cifar10_diff_var_system}, \ref{fig:cifar100_diff_var_system}, \ref{fig:adult_diff_var_system}, \ref{fig:purchase_diff_var_system}, and \ref{fig:texas_diff_var_system}, we can observe that as $\pi_{t} \rightarrow 1$, $U(\pi_{t})$ starts decreasing. This is because, as the \texttt{system}'s belief about the type of \texttt{end-user} approaches $1$, both the variance of the explanation generated by the system and the variance of the noisy explanation given to the \texttt{end-user} approach $u_{th}$ and $l_{th}$, respectively. After a certain point, i.e., when $ex^{v}_{s_{y}} > U(\pi_{t})$, the \texttt{system} will block the \texttt{end-user}, which confirms to our intuition.
%%when $ex_{ge}(\pi_{t}) > U(\pi_{t})$, the \texttt{system} will block the \texttt{end-user}.
\item From the optimal functional paths $L^{+}(\pi_{t})$,  $L^{-}(\pi_{t})$ and  $L(\pi_{t})$ of the \texttt{end-user} for each of the dataset, as shown in Figures \ref{fig:cifar10_diff_var_eu}, \ref{fig:cifar100_diff_var_eu}, \ref{fig:adult_diff_var_eu}, \ref{fig:purchase_diff_var_eu}, and \ref{fig:texas_diff_var_eu}, we can observe that as $ \pi_{t} \rightarrow 1$,  $L^{+}(\pi_{t})$,  $L^{-}(\pi_{t})$ and  $L(\pi_{t})$ approach the threshold $l_{th}$. As discussed in \ref{subsec:eqdesc}, as $ \pi_{t} \rightarrow 1$ and the variance of the explanation given to the \texttt{end-user} starts to approach the variance threshold, it means a malicious \texttt{end-user} is trying to compromise the \texttt{system}. Thus, if the \texttt{system} doesn't block the \texttt{end-user} at the right time (or doesn't have knowledge about optimal $U(\pi)$), then the \texttt{end-user} can easily compromise the \texttt{system}. 

\item Earlier we showed that a unique MPE exists when $L^{+}(\pi_{t})$ and $L^{-}(\pi_{t})$ begin to converge as $\pi_{t} \rightarrow 1$. 
%%So, we try to find out the first intersection point (root) of $L^{+}(\pi_{t})$ and $L^{-}(\pi_{t})$ as the unique MPE. 
This is also visible from our results as shown in Figure \ref{fig:baseline_game}, where we can observe that as $ \pi_{t} \rightarrow 1$, $L^{+}(\pi_{t})$ and  $L^{-}(\pi_{t})$ starts to converge. However, for the CIFAR-10 dataset, one can observe that the curves $L^{+}(\pi_{t})$ and  $L^{-}(\pi_{t})$ doesn't converge as $ \pi_{t} \rightarrow 1$. Thus, an MPE doesn't exist in the case of CIFAR-10 dataset. The intuition behind this observation is that the fluctuations (or variance) of the explanation variance computed for the CIFAR-10 is high, making it difficult for them to converge to a single point. 
%When curves $L^{+}(\pi_{t})$ and $L^{-}(\pi_{t})$ converge, $L(\pi_{t}) =L^{+}(\pi_{t})=L^{-}(\pi_{t})$ or we have reached an equilibrium. 
Finally, if the \texttt{system} doesn't block the \texttt{end-user} before the threshold $l_{th}$ or $u_{th}$ is reached, then we say a unique MPE exists in the game.  
\end{itemize}

\begin{figure}[!ht]
     \centering

      \begin{subfigure}[b]{0.49\linewidth}
         \centering
         \includegraphics[width=\textwidth,trim=0.2cm 0.4cm 0.2cm 0]{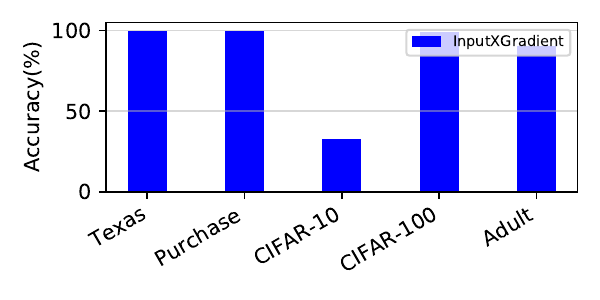}
         \caption{}
         \label{fig:acc_basemethod}
     \end{subfigure}
     \begin{subfigure}[b]{0.49\linewidth}
         \centering
         \includegraphics[width=\textwidth, trim=0.2cm 0.4cm 0.2cm 0]{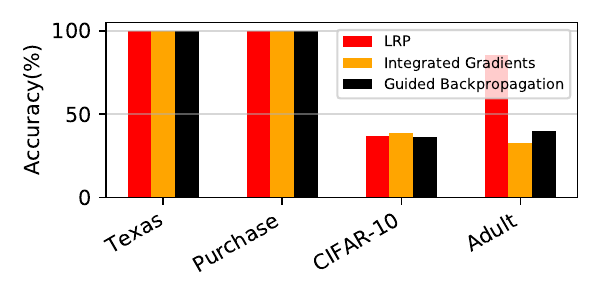}
         \caption{}
         \label{fig:acc_othermethod}
     \end{subfigure}\vspace{-0.15cm}
        \caption{Accuracy ($TPR$) for the optimal strategy obtained by the \texttt{system} and the \texttt{end-user}: a) $\emph{Gradient*Input}$ method and b) Other explanation methods.}
        \label{fig:base_acc}\vspace{-0.3cm}
\end{figure}

\noindent
\textbf{Attack Accuracy in the Baseline Setting: }After obtaining the optimal strategies, we use the range of the training data points of each dataset to determine how many data point variances lie below the computed threshold $u_{th}$ to determine their membership. As shown in Figure \ref{fig:acc_basemethod}, the attack accuracy for all the datasets except CIFAR-10 is more than 50\%. This result aligns with the observed game equilibrium analysis. Hence, the fluctuations in explanation variance make it difficult for an adversary to reach the target threshold, in consequence, to launch MIAs. From these obtained results, one can easily observe that the explanations provide a new opportunity or an attack vector to an adversary actively trying to compromise the \texttt{system}. In other words, our results are clear indicators that an adversary can repeatedly interact with the \texttt{system} to compute the explanation variance threshold and successfully launch membership inference attacks against the \texttt{system}.

\noindent
\textbf{Results for other Explanation Techniques: }
We also analyzed the game for the three other explanation methods considered in this paper. We do not plot the game evolution results in this setting as the plots follow a very similar trend as seen in Figure \ref{fig:baseline_game}, i.e., game equilibrium was achieved for all the datasets except for the CIFAR-10. We uses the same setting as the baseline setting (mentioned above) to compute attack accuracy for these three explanation methods.
%We again used training data points for each of the datasets: Texas, Purchase, CIFAR-10, and Adult, to compute the attack accuracy of an attacker.
%Thus, for the prediction label of any training data points, if the computed explanation variance is below $u_{th}$, then an attacker successfully determines the training membership of that data point.
We obtained each dataset's attack accuracy as shown in Figure \ref{fig:acc_othermethod}. %One can observe from this plot that 
For the Texas and Purchase datasets, 100\% accuracy was achieved, i.e., an attacker effectively determines the membership of all the data points used for training the model. However, for the CIFAR-10, the attack accuracy was below 50\%, and for the Adult dataset, attack accuracy was above 50\% only for the LRP explanation method. The reason is again the high fluctuations in the computed variance for the CIFAR-10 dataset (slightly less for the Adult dataset), thus making it difficult for an adversary to determine the membership of the data points in those datasets. These results clearly indicate that, for different explanation methods, 
%\mj{why sources? can we say ``methods"?},
an adversary's capability to launch MIA attacks will vary and may depend on the variance of the explanations.
%\mj{check this para - I made edits}
%\am{this paragraphs is poorly written compared to rest of the paper. i can improve this after kavita is done with MJ's comments}.

\iffalse
\begin{figure*}[H]
     \centering

      \begin{subfigure}[b]{0.2\textwidth}
         \centering
         \includegraphics[width=\textwidth,trim=0.3cm 0.4cm 0.3cm 0]{Figures/k_100/belief.pdf}
         \caption{}
         \label{fig:belief_100_old}
     \end{subfigure}
     \begin{subfigure}[b]{0.2\textwidth}
         \centering
         \includegraphics[width=\textwidth,trim=0.3cm 0.4cm 0.3cm 0]{Figures/k_100/belief_NM.pdf}
         \caption{}
         \label{fig:belief_100_new}
     \end{subfigure}
    \begin{subfigure}[b]{0.2\textwidth}
\centering
\includegraphics[width=\linewidth,trim=0.3cm 0.4cm 0.3cm 0]{Figures/k_100/ex_toeu_g_vs_tem_g_copy.pdf}
\caption{}
\label{fig:ex_toeu_vs_exgi}
\end{subfigure}\vspace{-0.15cm}
     
        \caption{Belief ($\pi_{t}$) evolution of the \texttt{system} for the model having $k=100$ classes and data points having $n_{f} = 100$ features.}
        \label{fig:base_acc}\vspace{-0.3cm}
\end{figure*}
\fi

%\vspace{-0.1cm}
\subsection{Analysis of other Relevant Factors}
\label{subsec:other}
In this section, we analyze the influence of other factors such as input dimension, underfitting, and overfitting, which can impact the equilibrium evolution, and thus the accuracy of the explanation-based threshold attacks. 
For this, we analyze the game's evolution on synthetic datasets and the extent to which an adversary can exploit model gradient information to conduct membership inference attacks. Using artificially generated datasets helps us identify the important factors that can facilitate information leaks.

\begin{itemize}[leftmargin=*]
\item \textbf{Impact of Input dimension. }
First, we analyze the impact of the input dimension of the data points on the game evolution. To generate datasets, we used the Sklearn make classification module \cite{scikit-learn}. The number of classes is set to either 2 or 100, while the number of features will vary in range $t_{f} \in [10, 100, 1000, 6000]$. For each setting, we sample 20,000 points and split them evenly into training and test set. Second, for each value $t_{f}$ and for each class, we employ two models to train from this data, namely, model $A$ and model $B$. Model $A$ is chosen to have less
%\am{fewer?}
layers (or depth) than model $B$; to compare the effect of the complexity of the models on the game evolution and attack accuracy. Model $A$ is a fully connected NN with two hidden layers
%\am{composed of}
fifty nodes each, the \textit{tanh} activation function between the layers, and \textit{softmax} as the final activation. The network is trained using \textit{Adagrad} with a learning rate of 0.01 and a learning rate decay of $10^{-7}$ for 50 epochs. Model $B$ is a five-layer fully connected NN with \textit{tanh} activations. The layer sizes are chosen as 2048, 1024, 512, 256 and 100. We use the \textit{Adagrad} optimizer with a learning rate of 0.01 and a learning rate decay of $10^{-7}$ to train the model for 50 epochs. Next, we demonstrate the effect of these models on our experiment's two main objectives. 
%\am{the two objectives are commented out earlier}.
%\K{Do you want me to write it down here?}

\begin{itemize}
    \item \textbf{Effect of Model $A$ on Game Evolution and Attack Accuracy. }For $k=2$ classes, we observe a similar trend in the game evolution as shown in Figure \ref{fig:baseline_game} for each of the features $t_{f} \in [10, 100, 1000, 6000]$. However, for $k=100$ classes, we observed that the belief $\pi_{t}$ of the \texttt{system} about the type of the \texttt{end-user} is always set to $1$ as shown in Figure \ref{fig:belief_old}. As a consequence, the variance of the explanations generated is equivalent to the variance of the explanations given, as shown in Figure \ref{fig:ex_toeu_vs_exgi}. Hence, the game didn't evolve as the \texttt{system} gave the same explanation to the \texttt{end-user}. The reason is because of underfitting. Model $A$ simply does not have enough depth (in terms of the number of layers) to accurately classify 100 classes. Thus, it was not able to accurately classify $100$ classes. Moreover, the loss of model $A$ was approximated to be around $5.74$ for all the features considered, which is not accurate. Hence, the model $A$ gives erroneous results for model predictions. In consequence, it impacts the two objectives of the experiments. 
    %\mj{instead of ``good" can we use the word ``accurate"?}.
\begin{figure}[H]
     \centering
     \vspace{-0.5cm}
      \begin{subfigure}[b]{0.49\linewidth}
         \centering
         \includegraphics[width=\textwidth, trim=0.2cm 0.4cm 0.2cm 0]{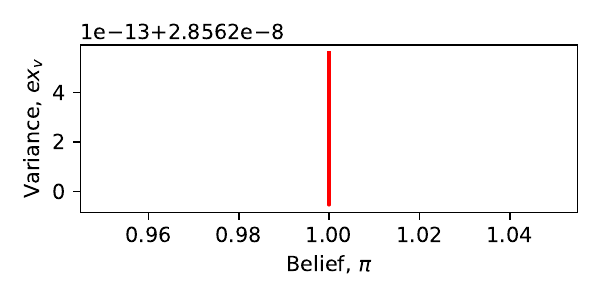}
         \caption{}
         \label{fig:belief_old}
     \end{subfigure}
     \begin{subfigure}[b]{0.49\linewidth}
         \centering
         \includegraphics[width=\textwidth, trim=0.2cm 0.4cm 0.2cm 0]{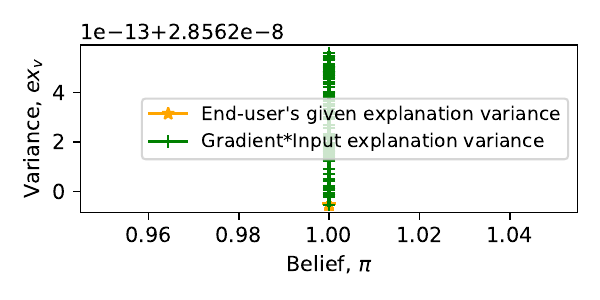}
         \caption{}
         \label{fig:ex_toeu_vs_exgi}
     \end{subfigure}\vspace{-0.3cm}
        \caption{Explanation variance generated vs. explanation variance given when $\pi_{t}=1$.}
    \label{fig:belief_pattern}\vspace{-0.3cm}
\end{figure} 
    \item \textbf{Effect of Model $B$ on Game Evolution and Attack Accuracy.} In this case, we considered model $B$ of higher complexity (more layers compared to model $A$) that led to different results compared to model $A$. For $k=2$ classes and $n_f = 10$, we did not observe any game evolution, because for the test data points explanation variance we got $\sigma > \mu$ and $\sigma > 1$. Consequently, the values of the computed future variance of the explanation, using Equation \ref{eqn:gbm}, were zero. However, for $t_{f} \in [100, 1000, 6000]$, we observed the game equilibrium and computed the attack accuracy by sampling from the training data points, as shown in Figure \ref{fig:k_2_nm}. For $t_{f}=1000$, an attack accuracy greater than 50\% was observed, however, for $t_{f}=100$ and $t_{f}=6000$, an attack accuracy less than 50\% was observed. For $k=100$ classes, we did not observe any equilibrium for any of the features $t_f$. Therefore, based on the last index value of the explanation variance simulated (i.e. at $t=100$), we computed the threshold $u_{th}$ and accordingly determined the attack accuracy for each of the features as shown in Figure \ref{fig:k_100_nm}. In this setting, the loss of the model $B$ was approximated to be around $0.8$ for all the features considered. 
\end{itemize}

The results obtained for models $A$ and $B$ indicate that the choice of the model used for the classification task on hand also affects the game evolution, and thus, an adversary's capability to launch MIA attacks against the \texttt{system}.

\iffalse
\begin{figure}[ht]
\centering
\includegraphics[width=0.75\linewidth]{Figures/k_100/ex_toeu_g_vs_tem_g_copy.pdf}
\caption{Explanation variance generated vs. explanation variance given when $\pi_{t}=1$.}
\label{fig:ex_toeu_vs_exgi}
\end{figure}
\fi

\begin{figure}[H]
     \centering

      \begin{subfigure}[b]{0.49\linewidth}
         \centering
         \includegraphics[width=\textwidth, trim=0.4cm 0.4cm 0.2cm 0]{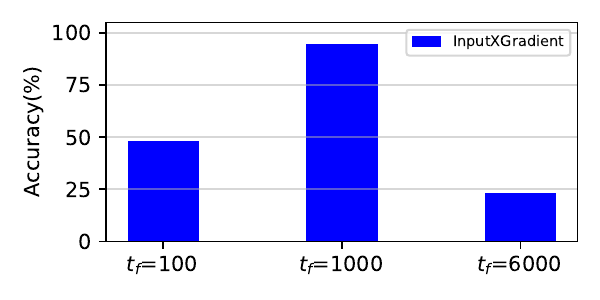}
         \caption{$k=2$.}
         \label{fig:k_2_nm}
     \end{subfigure}
     \begin{subfigure}[b]{0.49\linewidth}
         \centering
         \includegraphics[width=\textwidth, trim=0.4cm 0.4cm 0.2cm 0]{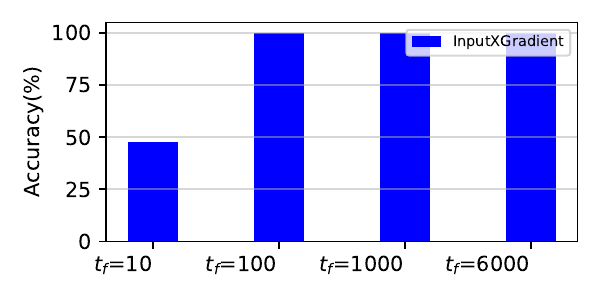}
         \caption{$k=100$.}
         \label{fig:k_100_nm}
     \end{subfigure}\vspace{-0.1cm}
        \caption{MIA accuracy for different features $n_f$ for model $B$.}
        \label{fig:k_nm}\vspace{-0.3cm}
\end{figure}

\item
\textbf{Impact of  Overfitting.} 
As detailed in \cite{yeom2018privacy}, overfitting can significantly increase the accuracy of the membership inference attacks. Thus, to see its effect on the game evolution and attack accuracy in our setup, we varied the number of epochs of the training process. We chose Purchase, Texas, and Adult datasets for this setting. As mentioned in Section \ref{sec:setup}, the models for these datasets are trained for 50 epochs; thus, to observe the impact of the number of epochs, we trained the models of each dataset for epochs in
%\mj{this sentence is not clear. What do you mean conducted experiments for epochs? Do you mean you trained for these many epochs?}
\{30, 50, 60\}.
%\mj{I don't think you should use square brackets here. Use paranthesis.e.g., \{30, 50, 60\}}.
The results are shown in Figure \ref{fig:epoch_acc}. We observed that overfitting only increased the attack accuracy for the Adult dataset, while it remained the same for the Texas dataset and decreased for the Purchased dataset. The reason is that the game evolution, and thus the MIA accuracy, depends on multiple factors (mentioned above), hence, the number of training epochs will not singularly determine or impact the capability of an adversary to launch MIA, as presented in the different scenarios mentioned earlier. In other words, the above results show that overfitting does not necessarily increase the attack's accuracy.

\begin{figure}[H]
\centering
\includegraphics[width=0.65\linewidth]{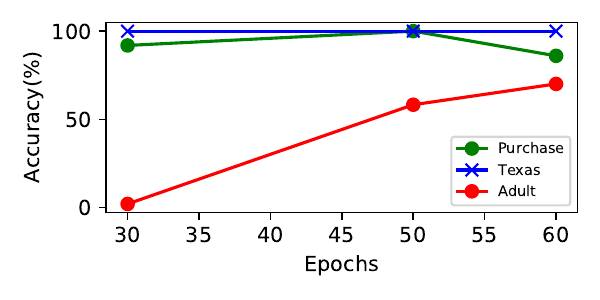}\vspace{-0.3cm}
\caption{MIA accuracy for different datasets trained for different epochs.}\vspace{-0.3cm}
\label{fig:epoch_acc}
\end{figure}

\end{itemize}

\section{Related Works}
\label{sec:related}

\changed{
In enhancing ML model transparency, privacy concerns have surfaced as a significant challenge. Existing literature highlights the susceptibility of ML models to privacy breaches through various attacks, including membership inference \cite{shokri2021privacy}, model reconstruction \cite{milli2019model, aivodji2020model}, model inversion \cite{zhao2021exploiting}, and sensitive attribute inference attacks 
 \cite{duddu2022inferring}. This paper specifically focuses on membership inference attacks or MIAs. MIA attack proposals in the literature can be broadly categorized into three main types based on the construction of the attack model: (i) \emph{binary classifier-based}, (ii) \emph{metric-based}, and (iii) \emph{differential comparison-based} approaches.

\begin{itemize}[leftmargin=*,noitemsep,nolistsep]
    \item \textbf{Binary Classifier-based Approaches: }Binary classifier-based MIAs involve training a classifier to distinguish the behavior of a target model's training members from non-members. The widely used shadow training technique \cite{shokri2017membership} is a prominent method, where an attacker creates shadow models to mimic the target model's behavior. This approach can be categorized as white box or black box, depending on the attacker's access to the target model's internal structure \cite{shokri2017membership, salem2018ml, yeom2018privacy, hui2021practical, nasr2019comprehensive, melis2019exploiting, leino2020stolen}.
    \item \textbf{Metric-based Approaches: }Metric-based MIAs simplify the process by calculating certain metrics on prediction vectors and comparing them to a preset threshold to determine the membership status of a data record. Categories within this approach include Prediction Correctness-based MIA \cite{long2017towards}(determines membership based on correct predictions), Prediction Loss-based MIA \cite{long2017towards}(infers membership based on prediction loss), Prediction Confidence-based MIA \cite{salem2018ml}(decides membership based on maximum prediction confidence), Prediction Entropy-based MIA \cite{yeom2018privacy}(uses prediction entropy to infer membership), and Modified Prediction Entropy-based MIA \cite{song2021systematic}(a modification addressing the limitation of existing prediction entropy-based MIA). 
    
    \item \textbf{Differential Comparison-based Approaches: }In differentially private models \cite{hui2021practical}, where the model preserves user privacy, a novel MIA named BLINDMI is proposed. In this work, BLINDMI generates a non-member dataset by transforming existing samples and differentially moves samples from a target dataset to this generated set iteratively. If the differential move increases the set distance, the sample is considered a non-member, and vice versa. %\commentM{This paragraph needs a bit of writing improvement. Why we talk about only one approach here?}.
\end{itemize}

While existing MIA approaches, such as binary classifier-based and metric-based techniques, have made notable contributions to understanding vulnerabilities in ML models, they come with inherent limitations. Binary classifier-based approaches may rely on assumptions about the adversary's knowledge of the target model's structure, potentially limiting their applicability in real-world scenarios where such information is not readily available. Metric-based approaches, while computationally less intensive, often face challenges in determining optimal thresholds for making membership inferences, leading to uncertainties regarding their accuracy. Additionally, these approaches may not provide theoretical guarantees for the optimality of the computed threshold. In contrast, our proposed approach focuses on the mathematical formulation of determining an optimal explanation variance threshold for adversaries to launch MIAs accurately.
%\commentM{Instead of ``100\%" shall we say ``perfect"? It would also be better to say something like ``...determining an optimal explanation variance threshold for adversaries to accurately launch MIAs."} accuracy.
By addressing the shortcomings of existing methods, our approach aims to offer a theoretically grounded and robust solution, enhancing the security of machine learning models against membership inference attacks.

Consequently, various defenses have been designed to counter MIAs, which fall into four main categories: confidence score masking \cite{yang2020defending, shokri2017membership, li2021membership, jia2019memguard}, regularization \cite{hayes2017logan, salem2018ml, nasr2019comprehensive, choquette2021label}, knowledge distillation \cite{shejwalkar2021membership, tang2022mitigating, zheng2021resisting}, and differential privacy \cite{chen2020gan, chen2018differentially, choquette2021label, hui2021practical, humphries2020differentially}. Confidence score masking, utilized in black-box MIAs on classification models, conceals true confidence scores, diminishing the effectiveness of MIAs. Regularization aims to reduce overfitting in target models, with methods such as Logan, ML, and label-based techniques to defend against MIAs. Knowledge distillation involves using a large teacher model to train a smaller student model, transferring knowledge to enhance defense against MIAs. Lastly, differential privacy (DP) is a privacy mechanism that offers an information-theoretical privacy guarantee when training ML models, preventing the learning or retaining of specific user details when the privacy budget is sufficiently small.

While existing defenses against Membership Inference Attacks (MIAs) have made valuable contributions, they have limitations. Although effective against black-box MIAs, confidence score masking may fall short in scenarios where sophisticated adversaries can adapt their strategies. Regularization techniques, while mitigating overfitting, might compromise the model's performance on the task on which it was trained. Knowledge distillation, reliant on a teacher-student model setup, introduces model architecture and training complexities. Differential privacy, though providing a robust privacy guarantee, often comes at the cost of reduced model utility due to noise injection during training. In contrast, our proposed approach, grounded in a mathematical formulation for the system to determine the optimal explanation variance threshold, aims to address these drawbacks by offering a theoretically sound and adaptable defense mechanism against MIAs.
}

%Next, as stated in the paper, we have
Game-theoretic approaches, such as zero-sum games \cite{dekel2010learning} \cite{globerson2006nightmare}, non-zero sum games \cite{dritsoula2017game}, sequential Bayesian games \cite{zhou2014adversarial} \cite{grosshans2013bayesian}, sequential Stackelberg games \cite{bruckner2011stackelberg} \cite{alfeld2017explicit} and simultaneous games \cite{dalvi2004adversarial} have been used in the research literature to model interactions with ML models, specifically to model adversarial classification. Contrary to these efforts, where an adversary's objective is to target the classification task of an ML model, our research effort focuses on the descriptive task, i.e., explaining the model predictions.
Specifically, we use a continuous-time stochastic Signaling Game \cite{neyman2017continuous} \cite{brazdil2009continuous} \cite{averboukh2016approximate} \cite{sobel2020signaling} to model the repeated interactions in a dynamic ML system with explanations to accomplish MIAs. 
We also make a novel use of GBM \cite{reddy2016simulating} \cite{hu1998optimal} \cite{dixit2012investment} to model the explanation variance in order to analyze how an adversary can utilize historical variance information to reach the target variance threshold.
To the best of our knowledge, there have been no prior works that utilize a continuous-time game-theoretic formulation to study the privacy leakages (in the form of MIAs) due to model explanations. Similar continuous-time stochastic signaling game models have been used in economic theory to study stock prices \cite{dixit2012investment}, dynamic limit pricing \cite{gryglewicz2009signaling} \cite{gryglewicz2019strategic}, and market trading \cite{daley2012waiting}. Our work is one of the first to use modeling concepts from economic theory to study the privacy problem in the ML and model explainability domain.
%of ML and explanation methods. 

%We used a continuous-time stochastic Signaling game \cite{neyman2017continuous} \cite{brazdil2009continuous} \cite{averboukh2016approximate} \cite{sobel2020signaling} to model such interactions. To the best of our knowledge, there has been no work that utilizes game theory to study the privacy leakages that can occur due to model explanations. Generally, continuous-time stochastic Signaling games are used in economic theory to study: stock prices \cite{dixit2012investment}, dynamic limit pricing \cite{gryglewicz2009signaling} \cite{gryglewicz2019strategic}, and market trading \cite{daley2012waiting}. We use GBM \cite{reddy2016simulating} \cite{hu1998optimal} \cite{dixit2012investment} to model the explanations variance to analyze how an adversary can utilize historical variance information to reach the target variance threshold. Thus, it is  the first work that integrates concepts from economics theory into the domain of ML and explanation methods to study MIAs. \vspace{-0.2cm}

\section{Conclusion}
\label{sec:conc}
We modeled the strategic interactions between an \texttt{end-user} and a \texttt{system}, where the variance of the explanations generated by the \texttt{system} evolve according to a stochastic differential equation, as a two-player continuous-time signaling game. Our main aim was to study how an adversary can launch explanation-based MIAs by utilizing the explanation variance threshold that he/she computes by repeatedly interacting with the \texttt{system}. In contrast to existing research efforts, this is the first work that analyzes the repeated interaction scenario, in which an adversary builds upon the previously gained knowledge to compute the optimal variance threshold. 
%We also conducted an extensive game evaluation; thus, the analysis of different gradient-based explanation methods and datasets in different settings demonstrates that: the capability of an adversary to launch MIA is dependent on multiple factors, and an adversary can effectively determine the membership of the training data points by utilizing the information on the variance of the explanations.
Our extensive experimental evaluation by considering different gradient-based explanation methods and publicly-available datasets in different settings demonstrates that the capability of an adversary to launch MIA is contingent upon multiple factors, such as the deployed explanation method, datapoint input dimension, model size, and model training rounds. Thus, a shrewd adversary with a working knowledge of all the aforementioned factors can effectively launch MIA by utilizing the information on the variance of the explanations.

\bibliographystyle{plain}
\bibliography{reference}

%\fi

%\appendix

\subsection{Lemma 1}
\label{app:lemma1}

\begin{proof}

If the variance of the explanation generated by the \texttt{system} reaches $u_{th}$, then \texttt{system's} belief about the type of \texttt{end-user} becomes zero, i.e., $\pi_{t} = 0$, as the \texttt{end-user's} type is revealed to be malicious. Let's assume that the \texttt{system} is perfectly able to block the \texttt{end-user} if he/she tries to compromise the \texttt{system}. Hence, using the \texttt{system's} value function, we get
    {
    \begin{equation}
    V(u_{th}) = \frac{K u_{th}}{r-\mu} - c_{d}  \label{eqn:one}
    \end{equation}
    \begin{equation}
    V^{'}(u_{th}) = \frac{K}{r-\mu}  \label{eqn:two}
    \end{equation}
    }
    
    Equation \eqref{eqn:one} is the value matching condition, which tells that a \texttt{system} will gain $k u_{th}$ while incurring a cost of $-c_{d}$ (detection cost) if it's able to block the \texttt{end-user} before he/she attacks. Equation \eqref{eqn:two} is a smooth pasting condition. 
    In Equation  \eqref{eqn:one} and  \eqref{eqn:two}, we replace $ V(u_{th})$ and it's derivative $V^{'}(u_{th})$ with the general solution of the value function of the \texttt{system} and it's derivative, respectively. As mentioned earlier, at $\pi = 0$ the explanation variance given by the \texttt{system} to the \texttt{end-user} becomes zero. As a result, since root $\beta_{2} < 0$, second term in the value function can go to $-\infty$. To eliminate this case we make the constant $B_{2}=0$  and we only calculate the constant $B_{1}$.Therefore, equations \eqref{eqn:one} and \eqref{eqn:two} are resolved to, 
    {
    $$B_{1} u_{th}^{\beta_{1}} + \frac{r_{e} u_{th}}{r-\mu} = \frac{k u_{th}}{r-\mu} - c_{d}$$
    $$B_{1} \beta_{1} u_{th}^{\beta_{1}-1} + \frac{r_{e}}{r-\mu} = \frac{k}{r-\mu}$$
    }
    
     We first compute the constant $B_{1}$, then find the threshold $u_{th}$. By solving the above two equations using the elimination and substitution method, we get
    
    {
    $$u_{th} = \frac{\beta_{1} c_{d} (r-\mu)}{(\beta_{1}-1) (k-r_{e})} > 0$$
    }
    
    As outlined in the payoff assumption, $k > r_e$, which implies that $u_{th} > 0$.
    \end{proof}

\subsection{Lemma 2}
\label{app:lemma2}
\begin{proof}
It is assumed that if an opportunity arrives for the \texttt{end-user} to compromise the \texttt{system}, then he/she will take one more step after signaling. This step can either make \texttt{end-user} to accomplish his/her's goal of compromising the \texttt{system} or get's detected and blocked by it. If \texttt{end-user} gets blocked by the \texttt{system}, the belief will jump to $\pi = 0$. Thus, to eliminate the possibility of the second term, $A_{2} l_{th}^{\beta_{2}} \to -\infty$ (as $\beta_{2}<0$) in the general solution of the \texttt{end-user's} value function, we assume $A_{2} = 0$.
    
Let's assume that \texttt{end-user} is successful in compromising the \texttt{system}, then the payoff gained by the \texttt{end-user} will be greater than the signaling stage payoff as there will be some monetary benefit to the \texttt{end-user} because of the privacy leak. Using the value matching and the smooth pasting conditions for \texttt{end-user} between signaling and not-signaling stages and accordingly assuming the payoffs received by the \texttt{end-user}, we get
    
    {
    \begin{equation}
    A_{1} l_{th}^{\beta_{1}} + \frac{P l_{th}}{r-\mu} < \frac{M_{NS}^{m} l_{th}}{r-\mu} - d \label{eqn:3}
    \end{equation}
    \begin{equation}
    A_{1} \beta_{1} l_{th}^{\beta_{1}-1} + \frac{P}{r-\mu} < \frac{M_{NS}^{m}}{r-\mu} \label{eqn:4}
    \end{equation}
    }
    Solving the above two equations we get, 
    {
    $$l_{th} > \frac{\beta_{1} d (r-\mu)}{(\beta_{1}-1) (M_{NS}^{m}-P)} > 0$$
    }
    $M_{NS}^{m} = M^{m} + d^{'}$, if \texttt{end-user} is successful in compromising the \texttt{system}. Thus, $M_{NS}^{m} > P$, which implies $l_{th} > 0$.

\end{proof}

\subsection{$L^{+}(\pi)$ and $L^{-}(z)$ conditions}
\label{app:conditions}
Now, we will define the conditions under which an optimal explanation variance path $L(\pi)$ (given optimal $U(\pi)$) exists for the \texttt{end-user}. As mentioned earlier, we are trying to find an MPE in a pooling situation, which will occur when \texttt{end-user's} explanation lies in the range, $(0, l_{th})$. $l_{th}$ is the explanation variance threshold, which an \texttt{end-user} is trying to achieve to accomplish his/her aim of compromising the \texttt{system}. To find optimal $L(\pi)$, we define two increasing functions, $L^{+}(\pi)$ and $L^{-}(\pi)$. First, we will prove that these two functions exist and then show these two converge by finding out the first intersection point of these two curves as the root of a polynomial (defined below) and finally show that a unique root of that polynomial exists. We will write 3 equations each for $L^{+}(\pi)$ and $L^{-}(z)$ according to the value matching, smooth pasting, and the condition in which the variance of the explanation received is opposite of what \texttt{end-user} expected. That is, if instead of $L^{+}(\pi)$, an \texttt{end-user} gets $L^{-}(\pi)$ or if  instead of $L^{-}(\pi)$, an \texttt{end-user} gets $L^{+}(\pi)$ ).
    Three conditions for $L^{+}(\pi)$ are: -
    {
    $$F(L^{+}(\pi), \pi) = \lambda(L^{+}(\pi), \pi)$$
}
Replacing $F(L^{+}(\pi), \pi)$ with it's solution, we get
        \begin{equation}
    A_{1}^{+}(\pi) L^{+}(\pi)^{\beta_{1}} + A_{2}^{+}(\pi)L^{+}(\pi)^{\beta_{2}} \nonumber
     \end{equation}
     {
    \begin{equation}
       + \frac{P L^{+}(\pi)}{r-\mu} = \lambda(L^{+}(\pi), \pi) \label{eqn:11}
    \end{equation}
   
    }

    {
    $$F^{'}_{L^{+}}(L^{+}(\pi), \pi) = \lambda^{'}(L^{+}(\pi), \pi)$$
    }
    Replacing $F^{'}_{L^{+}}(L^{+}(\pi), \pi)$ with it's solution, we get
    {
    $$ A_{1}^{+}(\pi) \beta_{1} L^{+}(\pi)^{\beta_{1}-1} + A_{2}^{+}(\pi) \beta_{2} L^{+}(\pi)^{\beta_{2}-1} $$
    \begin{equation}
       + \frac{P }{r-\mu} = \lambda^{'}(L^{+}(\pi), \pi) \label{eqn:12}
    \end{equation}
    }

    {
    $$F(L^{-}(\pi), \pi) = \frac{P \cdot L^{-}(\pi)}{r-\mu}$$
    }
     Replacing $F(L^{-}(\pi), \pi) $ with it's solution, we get
    \begin{equation}
        A_{1}^{+}(\pi) L^{-}(\pi)^{\beta_{1}} + A_{2}^{+}(\pi)L^{-}(\pi)^{\beta_{2}} = 0 \label{eqn:13}
    \end{equation}
    
    The above conditions are at the boundary between signaling and not-signaling. Equation \eqref{eqn:11} states that whether successful in compromising the \texttt{system} or not, \texttt{end-user} will get $\lambda(L^{+}(\pi), \pi)$ given the state of the game $(ex, \pi)$. Equation \eqref{eqn:12} is the smooth pasting condition and equation \eqref{eqn:13} is when instead of getting $L^{+}(\pi)$, \texttt{end-user} gets $L^{-}(\pi)$. Thus, will receive ($P \cdot L^{-}(\pi)$) from the \texttt{system}.

    Three conditions for $L^{-}(\pi)$ are: -
    {
    $$F(L^{-}(\pi), \pi) = \frac{P L^{-}(\pi)}{r-\mu}$$
    
 Replacing $F(L^{-}(\pi), \pi) $ with it's solution, we get
    \begin{equation}
        A_{1}^{-}(\pi) L^{-}(\pi)^{\beta_{1}} + A_{2}^{-}(\pi)L^{-}(\pi)^{\beta_{2}} = 0 \label{eqn:14}
    \end{equation}
    }
    
    {
    $$F^{'}_{L^{-}}(L^{-}(\pi), \pi) = 0$$
     Replacing $F^{'}_{L^{-}}(L^{-}(\pi), \pi) $ with it's solution, we get
    \begin{equation}
        A_{1}^{-}(\pi) \beta_{1} L^{-}(\pi)^{\beta_{1}-1} + A_{2}^{-}(\pi) \beta_{2} L^{-}(\pi)^{\beta_{2}-1} = 0 \label{eqn:15}
    \end{equation}
    }
    
    {
    $$F(L^{+}(\pi), \pi) = \lambda(L^{+}(\pi), \pi)$$
    
  Replacing $F(L^{+}(\pi), \pi) $ with it's solution, we get
    \begin{equation}
        A_{1}^{-}(\pi) L^{+}(\pi)^{\beta_{1}} + A_{2}^{-}(\pi)L^{+}(\pi)^{\beta_{2}} + \frac{P L^{+}(\pi)}{r-\mu} = \lambda(L^{+}(\pi), \pi) \label{eqn:16}
    \end{equation}
    }
    
    Equation \eqref{eqn:14} is the value matching condition which states that if the \texttt{end-user} takes one more step, and gets at least the payoff one was getting in the previous stage, then it will stop playing the game. The reason is that $L^{-}(\pi)$ is a value with the highest amount of noise being added to it, hence \texttt{end-user} is already not getting anything. Thus, will want to stop attacking the \texttt{system}. Equation \eqref{eqn:15} is a smooth pasting condition and equation \eqref{eqn:16} is when \texttt{end-user} gets $L^{+}(\pi)$ instead of $L^{-}(\pi)$. Thus, there is a chance that \texttt{end-user} can succeed in compromising the \texttt{system}. Hence, getting $\lambda(L^{+}(\pi), \pi)$.
    %In appendix \ref{sec:lemma3} and \ref{sec:lemma4}, we have shown the continuity and differentiability of $L^{+}(\pi)$ and $L^{-}(\pi)$ respectively. 
    The result from these two lemmas will be used in proving the existence and uniqueness of MPE in the game. 

\subsection{Lemma 3}
\label{app:lemma3}
    \begin{proof}
    To compute optimal  $L^{+}(\pi)$, we assume $\pi$ is set to 1. The reason being, since it's the variance of an explanation with maximum relevance, the \texttt{system's} belief about the type of \texttt{end-user} is set to 1.
    Given belief $\pi=1$, we solve equation \eqref{eqn:14} and \eqref{eqn:15} by substitution and elimination approach to get constants $A_{1}^{-}(\pi)$ and $A_{2}^{-}(\pi)$.
   
%   \begin{equation}
 %      A_{1}^{-}(\pi) = \frac{sig (1-\beta_{2})}{(r-\mu) (\beta_{2} - \beta_{1}) L^{-}(z)^{\beta_{1}-1}} + \frac{\beta_{2} V_{ex}}{(\beta_{2} - \beta_{1}) L^{-}(z)^{\beta_{1}}}
  % \end{equation}
   
   %\begin{equation}
    %   A_{2}^{-}(z) = \frac{sig (1-\beta_{1})}{(r-\mu) (\beta_{1} - \beta_{2}) L^{-}(z)^{\beta_{2}-1}} + \frac{\beta_{1} V_{ex}}{(\beta_{1} - \beta_{2}) L^{-}(z)^{\beta_{2}}}
   %\end{equation}
   
   We get $A_{1}^{-}(\pi) = 0$ and $A_{2}^{-}(\pi) = 0$.

 %  Now, we will calculate $\frac{\partial A_{1}^{-}(z)}{\partial z}$ and $\frac{\partial A_{2}^{-}(z)}{\partial z}$ to check for the continuity and differentiability of $A_{1}^{-}(z)$ and $A_{2}^{-}(z)$.
   
  % \begin{equation}
   %    \frac{\partial A_{1}^{-}(z)}{\partial z} = \frac{sig (1-\beta_{2})}{(r-\mu) (\beta_{2} - \beta_{1})} - \frac{\partial L^{-}(z)}{\partial z} \times \frac{\beta_{1}}{L^{-}(z)} \times A_{1}^{-}(z)
   %\end{equation}
   
   %\begin{equation}
    %   \frac{\partial A_{2}^{-}(z)}{\partial z} = \frac{sig (1-\beta_{1})}{(r-\mu) (\beta_{1} - \beta_{2})} - \frac{\partial L^{-}(z)}{\partial z} \times \frac{\beta_{2}}{L^{-}(z)} \times A_{2}^{-}(z)
  % \end{equation}
    
    Next, using boundary condition \eqref{eqn:16} and substituting the values of the constants $A_{1}^{-}(\pi)$ and $A_{1}^{-}(\pi)$ in this equation, we get an implicit expression for $L^{+}(\pi)$,
    {
    \begin{equation}
        \frac{P L^{+}(\pi)}{r-\mu} = \lambda(L^{+}(\pi), \pi) \nonumber 
    \end{equation}
    }
    
    By simplifying the above equation, we get
    {
    \begin{equation}
        L^{+}(\pi) = \frac{\lambda(L^{+}(\pi), \pi) (r-\mu)}{P} \label{eqn:17}
    \end{equation}
    }
    
    Taking the derivative of equation 25) we get, 
    {
    \begin{equation}
        \frac{\partial L^{+}(\pi)}{\partial z} = \frac{\lambda^{'}(L^{+}(\pi), \pi) (r-\mu)}{P} \label{eqn:18}
    \end{equation}
    }
    
    Since $\lambda^{'}(L^{+}(\pi), \pi) > 0$, $ L^{+}(\pi)$ is a well-defined, increasing, continuous and differentiable function. Hence proved.
    
    \end{proof}

\subsection{Lemma 4}
\label{app:lemma4}
 
   \begin{proof}

   Given belief $\pi$, we solve equations \eqref{eqn:11} and  \eqref{eqn:12}  by substitution and elimination approach to get $A_{1}^{+}(\pi)$ and $A_{2}^{+}(\pi)$, 
   
   {
   $$A_{1}^{+}(\pi) = \frac{\beta_{2} \lambda(L^{+}(\pi), \pi) - \lambda^{'}(L^{+}(\pi), \pi) L^{+}(\pi)} {(\beta_{2} - \beta_{1}) L^{+}(\pi)^{\beta_{1}}}$$
   \begin{equation}
        +\frac{P (1-\beta_{2})}{(r-\mu) (\beta_{2} - \beta_{1}) L^{+}(\pi)^{\beta_{1}-1}}  \label{eqn:19}
   \end{equation}
  } 
  {
  $$A_{2}^{+}(\pi) = \frac{\beta_{1} \lambda(L^{+}(\pi), \pi) - \lambda^{'}(L^{+}(\pi), \pi) L^{+}(\pi)} {(\beta_{1} - \beta_{2}) L^{+}(\pi)^{\beta_{2}}} $$
   \begin{equation}
       +\frac{P (1-\beta_{1})}{(r-\mu) (\beta_{1} - \beta_{2}) L^{+}(\pi)^{\beta_{2}-1}}  \label{eqn:20}
   \end{equation}
   }
   
   Now, we will calculate $\frac{\partial A_{1}^{+}(\pi)}{\partial z}$ and $\frac{\partial A_{2}^{+}(\pi)}{\partial z}$ to check for the continuity $A_{1}^{+}(\pi)$ and $A_{2}^{+}(\pi)$ which we will use in checking the continuity of $L^{-}(\pi)$.
   
   {
   $$\frac{\partial A_{1}^{+}(\pi)}{\partial \pi} = \frac{1}{\beta_{2} - \beta_{1}} \times \frac{\partial L^{+}(\pi)}{\partial z} (\frac{\beta_{2} \lambda^{'}(L^{+}(\pi), \pi)}{L^{+}(\pi)^{\beta_{1}}} $$
   $$- \frac{\lambda^{''}(L^{+}(\pi), \pi)}{L^{+}(\pi)^{\beta_{1} -1}})  + \frac{\partial L^{+}(\pi)}{\partial \pi} \times \frac{1}{L^{+}(\pi)} (A_{1}^{+}(\pi) (1-\beta_{1}) $$
   \begin{equation}
       - \frac{\beta_{2}\lambda(L^{+}(\pi), \pi)}{(\beta_{2} - \beta_{1})L^{+}(\pi)^{\beta_{1}}})  \label{eqn:21}
   \end{equation}
   }
   
   {
   $$\frac{\partial A_{2}^{+}(\pi)}{\partial z} = \frac{1}{\beta_{1} - \beta_{2}} \times \frac{\partial L^{+}(\pi)}{\partial z} (\frac{\beta_{1} \lambda^{'}(L^{+}(\pi), \pi)}{L^{+}(\pi)^{\beta_{2}}} $$
   $$- \frac{\lambda^{''}(L^{+}(\pi), \pi)}{L^{+}(\pi)^{\beta_{2} -1}})+ \frac{\partial L^{+}(\pi)}{\partial \pi} \times \frac{1}{L^{+}(\pi)} (A_{1}^{+}(\pi) (1-\beta_{2}) $$
   \begin{equation}
        - \frac{\beta_{1}\lambda(L^{+}(\pi), \pi)}{(\beta_{1} - \beta_{2})L^{+}(\pi)^{\beta_{2}}})  \label{eqn:22}
   \end{equation}
   }
   Using boundary condition \eqref{eqn:13} and taking the derivative of it, gives an implicit expression for $\frac{\partial L^{-}(\pi)}{\partial \pi}$,
   {
    $$\frac{\partial A_{1}^{+}(\pi)}{\partial z} L^{-}(\pi)^{\beta_{1}} + \frac{\partial A_{2}^{+}(\pi)}{\partial z} L^{-}(\pi)^{\beta_{2}}$$
    $$+ A_{1}^{+}(\pi) \beta_{1} L^{-}(\pi)^{\beta_{1}-1} \frac{\partial L^{-}(\pi)}{\partial z} $$
    \begin{equation}
    + A_{2}^{+}(\pi) \beta_{2} L^{-}(\pi)^{\beta_{2}-1} \frac{\partial L^{-}(\pi)}{\partial \pi}  \label{eqn:23}
    \end{equation}
    }
   
   By simplifying the above equation, we get
   {
   \begin{equation}
       \frac{\partial L^{-}(\pi)}{\partial \pi} = - \frac{\frac{\partial A_{1}^{+}(\pi)}{\partial \pi} L^{-}(\pi)^{\beta_{1}} + \frac{\partial A_{2}^{+}(\pi)}{\partial \pi} L^{-}(\pi)^{\beta_{2}}}{A_{1}^{+}(\pi) \beta_{1} L^{-}(\pi)^{\beta_{1}-1} + A_{2}^{+}(\pi) \beta_{2} L^{-}(\pi)^{\beta_{2}-1}}  \label{eqn:24}
   \end{equation}
   }
   
   For equation \eqref{eqn:24} to be greater than zero, either the numerator (represented as $n(L^{-}(\pi))$) or the denominator (represented as $d(L^{-}(\pi))$) has to be negative. 
   
   \begin{equation}
       n(L^{-}(\pi)) = \frac{\partial A_{1}^{+}(z)}{\partial \pi} L^{-}(\pi)^{\beta_{1}} + \frac{\partial A_{2}^{+}(\pi)}{\partial \pi} L^{-}(\pi)^{\beta_{2}}  \label{eqn:25}
   \end{equation}
   
   \begin{equation}
       d(L^{-}(\pi)) = A_{1}^{+}(\pi) \beta_{1} L^{-}(\pi)^{\beta_{1}-1} + A_{2}^{+}(\pi) \beta_{2} L^{-}(\pi)^{\beta_{2}-1}  \label{eqn:26}
   \end{equation}
   
   Thus, $\frac{\partial L^{-}(\pi)}{\partial \pi} > 0$ or $L^{-}(\pi) > 0$ if and only if either $n(L^{-}(\pi)) < 0$ or $ d(L^{-}(\pi))<0$. 
   
  % [Conditions for which $\frac{\partial L^{-}(\pi)}{\partial \pi} > 0$ is true are dependent on many variables, that's why I haven't mentioned each of them. I can write it later if you want me to.]
      
   \end{proof}

\subsection{Theorem 1}
\label{app:theorem1}

\begin{proof}
    We are assuming optimal cutoff function, $U(\pi_{t})$, of the \texttt{system} is optimal as $U(\pi_{t})$ is computed using the belief it has about the type of the \texttt{end-user}. Using three boundary conditions:- value matching, smooth pasting, and instead of $U(\pi_{t})$, the \texttt{system} computes $L(\pi_{t})$ given belief $\pi_{t}$, we calculate the optimal $U(\pi_{t})$. These conditions are assumed at the boundary when the \texttt{end-user} has deviated, and the \texttt{system} has successfully detected the attack. In consequence, account of that \texttt{end-user} is suspended. When \texttt{system} computes $L(\pi_{t})$, which is below $U(\pi_{t})$, it will get a payoff of $r_{e} L(\pi_{t})$. The reason being, since \texttt{system} computes $L(\pi_{t})$, \texttt{end-user} has no incentive to deviate, thus will keep imitating the honest user. As a result, \texttt{system} will get a signaling stage payoff. The three boundary conditions are:
    {
    $$V(U(\pi_{t}),z) = \frac{k U(\pi_{t})}{r-\mu} - c_{d}$$
    $$V^{'}_{U}(U(\pi_{t}), \pi) = \frac{k}{r-\mu}$$
    $$V(L(\pi_{t}), \pi) = \frac{r_{e} L(\pi_{t})}{r-\mu}$$
    }
    
    By replacing the value of $V(U(\pi_{t}), \pi)$, $V(L(\pi_{t}), \pi)$ and $V^{'}_{U}(U(\pi_{t}), \pi)$ with the general solution of the value function of the \texttt{system} and it's derivative, we get the following equations:-
    {\scriptsize
    \begin{equation}
      B_{1}(\pi) U(\pi_{t})^{\beta_{1}} + B_{2}(z) U(\pi_{t})^{\beta_{2}} + \frac{r_{e} U(\pi_{t})}{r-\mu}  = \frac{k U(\pi_{t})}{r-\mu} - c_{d}  \label{eqn:5}
    \end{equation}
}
{\scriptsize
    \begin{equation}
      B_{1}(\pi) \beta U(\pi_{t})^{\beta_{1} -1} + B_{2}(\pi) \beta U(\pi_{t})^{\beta_{2}-1} + \frac{r_{e}}{r-\mu} = \frac{k}{r-\mu} \label{eqn:6}
    \end{equation}
}
{\scriptsize
    \begin{equation}
         B_{1}(z) L(\pi_{t})^{\beta_{1}} + B_{2}(\pi) L(\pi_{t})^{\beta_{2}} + \frac{r_{e} L(\pi_{t})}{r-\mu}  = \frac{r_{e} L(\pi_{t})}{r-\mu} \label{eqn:7}
    \end{equation}
}    

    For simplicity, we will be replacing $L(\pi_{t})$ with $L(\pi)$ and $U(\pi_{t})$ with $U(\pi)$.
    Solving equations \eqref{eqn:5} and  \eqref{eqn:6} using substitution and elimination, we get the coefficients $B_{1}$ and $B_{2}$.
    {\scriptsize
    $$B_{1} = \frac{(\beta_{2}-1) (k-r_{e})}{(r-\mu) (\beta_{2}-\beta_{1}) U(\pi)^{\beta_{1}-1}} - \frac{c_{d} \beta_{2}}{(\beta_{2}-\beta_{1}) U(\pi)^{\beta_{1}}}$$
    }
    
    {\scriptsize
    $$B_{2} = \frac{(\beta_{1}-1) (k-r_{e})}{(r-\mu) (\beta_{1}-\beta_{2}) U(\pi)^{\beta_{2}-1}} - \frac{c_{d} \beta_{1}}{(\beta_{1}-\beta_{2}) U(\pi)^{\beta_{2}}}$$
    }
    
    Now, in order to compute the optimal $U(\pi)$, we substitute the values of $B_1$ and $B_2$ computed above in equation  \eqref{eqn:7}:
    
    {\scriptsize
    $$(\frac{(\beta_{2}-1) (k-r_{e})}{(r-\mu) (\beta_{2}-\beta_{1}) U(\pi)^{\beta_{1}-1}} - \frac{c_{d} \beta_{2}}{(\beta_{2}-\beta_{1}) U(\pi)^{\beta_{1}}}) L(\pi)^{\beta_{1}}$$
    $$(\frac{(\beta_{1}-1) (k-r_{e})}{(r-\mu) (\beta_{1}-\beta_{2}) U(\pi)^{\beta_{2}-1}} - \frac{c_{d} \beta_{1}}{(\beta_{1}-\beta_{2}) U(\pi)^{\beta_{2}}}) L(\pi)^{\beta_{2}}=0$$
    }
    
    By simplifying the above equation, we get
    
    {\scriptsize
    $$\frac{(k-r_{e}) U(\pi)}{(r-\mu) (\beta_{2}-\beta_{1})} [ (\beta_{2}-1) J(\pi, t)^{\beta_{1}} - (\beta_{1}-1) J(\pi, t)^{\beta_{2}}] $$
    $$- \frac{c_{d}}{\beta_{2}-\beta_{1}} [\beta_{2} J(\pi, t)^{\beta_{1}} - \beta_{1} J(\pi, t)^{\beta_{2}}] = 0$$
    }
    
    {\scriptsize
    \begin{equation}
        U(\pi) = \frac{\left[\beta_{2} J(\pi, t)^{\beta_{1}} - \beta_{1} J(\pi, t)^{\beta_{2}}\right] \left( r-\mu\right)}{\left[ (\beta_{2}-1) J(\pi, t)^{\beta_{1}} - (\beta_{1}-1) J(\pi, t)^{\beta_{2}}\right] \left( k-r_{e}\right)} \label{eqn:8}
    \end{equation}
    }
    
    Equation  \eqref{eqn:8} allows us to define a "best response" threshold curve $U(\pi)$ for the \texttt{system}. We use $J(\pi, t)$ to denote the value of $\frac{L(\pi)}{U(\pi)}$. Now, we will calculate the derivative $\frac{\partial U(\pi)}{\partial \pi}$ to check whether $U(\pi)$ is non-increasing or not. The derivative of $U(\pi)$ depends on it's numerator (represented as $n(U(\pi))$) and denominator (represented as $d(U(\pi))$) and are given as:
    
    {
    $$
        n(U(\pi)) = \left[\beta_{2} J(\pi, t)^{\beta_{1}} - \beta_{1} J(\pi, t)^{\beta_{2}}\right] \left( r-\mu\right)
    $$
    }
    {
    $$
        d(U(\pi)) = \left[ (\beta_{2}-1) J(\pi, t)^{\beta_{1}} - (\beta_{1}-1) J(\pi, t)^{\beta_{2}}\right] \left( k-r_{e}\right)
    $$
    }

    The derivatives of $n(U(\pi))$ and $d(U(\pi))$ are given as:
    {\scriptsize
    \begin{equation}
       n_{U}(U(\pi)) = [\beta_{1}\beta_{2} J^{`}(\pi, t)^{\beta_{1}-1} - \beta_{2}\beta_{1} J^{`}(\pi, t)^{\beta_{2}-1}] \times \left( r-\mu\right) \label{eqn:9}
    \end{equation}
}

    {\scriptsize
    \begin{equation}
      d_{U}(U(\pi)) = [ \beta_{1}(\beta_{2}-1) J^{`}(\pi, t)^{\beta_{1}-1}   - \beta_{2}(\beta_{1}-1) J^{`}(\pi, t)^{\beta_{2}-1}] \times \left( k-r_{e}\right) \label{eqn:10}
    \end{equation}
    }
    
    Thus, $U(\pi)$ will be non-increasing, if either $n_{U}(U(\pi)) < 0$ or $d_{U}(U(\pi)) < 0$.

    \textit{Checking when $n_{U}(U(\pi)) > 0$: }As we know $\beta_{1} > 1$, $\beta_{2} < 0$ and $r-\mu > 0$. Thus, $n_{U}(U(\pi)) > 0$ if and only if $\beta_{2}\beta_{1} J^{`}(\pi, t)^{\beta_{2}-1} > \beta_{1}\beta_{2} J^{`}(\pi, t)^{\beta_{1}-1}$ and $n_{U}(U(\pi)) < 0$ otherwise.
    
    \textit{Checking when $d_{U}(U(\pi) > 0$: ) }We know $k-r_{e} > 0$. Thus, $d_{U}(U(\pi)) > 0$ if and only if $\beta_{2}(\beta_{1}-1) J^{`}(\pi, t)^{\beta_{2}-1} > \beta_{1}(\beta_{2}-1) J^{`}(\pi, t)^{\beta_{1}-1}$ and $n_{U}(U(\pi)) < 0$ otherwise.
    
    So, $U(\pi)$ will be a non-increasing function if and only if the above conditions are satisfied pertaining to either $n_{U}(U(\pi)) < 0$ or $d_{U}(U(\pi) < 0$.
    
    \end{proof}

\subsection{Theorem 2}
\label{app:theorem2}
    \begin{proof}
    Computation of $L(\pi)$ follows the same approach as the computation of $L^{+}(\pi)$, except the belief is not set to 1, i.e., $\pi \neq 1$. The reason being it's an optimal functional path for the \texttt{end-user} computed at each belief ($\pi$) of the \texttt{system} about the type of the \texttt{end-user}. So, $L(\pi)$ is formulated as:
    
     {
    \begin{equation}
        L(\pi) = \frac{\lambda(L(\pi), \pi) (r-\mu)}{P} \nonumber
    \end{equation}
    }
    
    Taking the derivative of the above equation we get, 
    {
    \begin{equation}
        \frac{\partial L(\pi)}{\partial z} = \frac{\lambda^{'}(L(\pi), \pi) (r-\mu)}{P} \nonumber
    \end{equation}
    }
    
    Since $\lambda^{'}(L(\pi), \pi) > 0$ and $P>0$, $ L(\pi)$ is a well-defined, increasing, continuous and differentiable function. Hence proved.
    \end{proof}

\subsection{Theorem 3}
\label{app:theorem3}
\begin{proof}
In this proof, first, we will convert the system of equations (\eqref{eqn:11}, \eqref{eqn:12}, \eqref{eqn:13}, \eqref{eqn:14}, \eqref{eqn:15}, \eqref{eqn:16}) into a system of two equations (shown below). Then, we will try to prove that a unique root exists for this pair of equations. Hence, any solution to this pair of equations will also be a root of the system of equations (\eqref{eqn:11}, \eqref{eqn:12}, \eqref{eqn:13}, \eqref{eqn:14}, \eqref{eqn:15}, \eqref{eqn:16}). If this root exists and is unique, we say a a unique MPE exists in the game. The pair of equations are derived as follows:  

First, we replace the constants in equations \eqref{eqn:13} and \eqref{eqn:16}, then after rearranging the terms, we get the following two equations:  

{\scriptsize
$$ \left[\frac{\beta_{2} \lambda(L^{+}(\pi), \pi) - \lambda^{'}(L^{+}(\pi), \pi) L^{+}(\pi)} {(\beta_{2} - \beta_{1})} +\frac{P (1-\beta_{2}) L^{+}(\pi)}{(r-\mu) (\beta_{2} - \beta_{1})} \right] \times (\frac{1}{\varsigma})^{\beta_{1}}$$
}
{\scriptsize
\begin{equation}
+ \left[\frac{\beta_{1} \lambda(L^{+}(\pi), \pi) - \lambda^{'}(L^{+}(\pi), \pi) L^{+}(\pi)} {(\beta_{1} - \beta_{2})} +\frac{P (1-\beta_{1}) L^{+}(\pi)}{(r-\mu) (\beta_{1} - \beta_{2})} \right] \times (\frac{1}{\varsigma})^{\beta_{2}} \label{eqn:27}
\end{equation}
 }

{\scriptsize
\begin{equation}
    \frac{P \varsigma}{r-\mu} - \frac{\lambda(L^{+}(\pi), \pi)}{L^{-}(\pi)} = 0 \label{eqn:28}
\end{equation}
}

In the above equations, we have introduced a term $\varsigma$, which is equal to $\frac{L^{+}(\pi)}{L^{-}(\pi)}$ and is a new dependent variable. Thus, any possible root of the above pair of equations will be a value of $\frac{L^{+}(\pi)}{L^{-}(\pi)}$ or of $\varsigma$.

\noindent
\textbf{Proof of the existence of $\varsigma$: }
We will be using concepts from the calculus to prove that a root exists for equations \eqref{eqn:27} and \eqref{eqn:28}.

From equation \eqref{eqn:28}, we get one of the value of $\varsigma$,
{\scriptsize
\begin{equation}
    \varsigma = \frac{\lambda(L^{+}(\pi), \pi) \times (r-\mu)}{P \times L^{-}(\pi)} \label{eqn:29}
\end{equation}
}

To show that this is the root for the system of equations \eqref{eqn:27} and \eqref{eqn:28}, we take the curve represented by equation \eqref{eqn:27} and denote it by $\Lambda$. Then, we will show that only one real root of $\Lambda$ exists in the range $ex \in [0, l_{th}]$, which is the solution to the six boundary conditions.

{\scriptsize
\begin{equation}
\begin{split}
\Lambda (\varsigma) = \left[\frac{\beta_{2} \lambda(L^{+}(z), z) - \lambda^{'}(L^{+}(z), z) L^{+}(z)} {(\beta_{2} - \beta_{1})} +\frac{P (1-\beta_{2}) L^{+}(z)}{(r-\mu) (\beta_{2} - \beta_{1})} \right] \\ \times (\frac{1}{\varsigma})^{\beta_{1}} \\ + \left[\frac{\beta_{1} \lambda(L^{+}(z), z) - \lambda^{'}(L^{+}(z), z) L^{+}(z)} {(\beta_{1} - \beta_{2})} +\frac{P (1-\beta_{1}) L^{+}(z)}{(r-\mu) (\beta_{1} - \beta_{2})} \right] \\ \times (\frac{1}{\varsigma})^{\beta_{2}} 
\end{split}
\label{eqn:30}
\end{equation}
}

The minimum explanation which can be given by the \texttt{system} to the \texttt{end-user} is zero which is at $\pi=0$, thus at this belief, both $L^{-}(\pi) \to 0$ and $L^{+}(\pi) \to 0$. As a result of which, $\varsigma = \frac{0}{0}$. To remove this case, we will assume that $L^{-}(\pi)$ is close to zero, not exactly 0. For the sake of simplicity, let's denote the coefficient of $(\frac{1}{\varsigma})^{\beta_{1}}$ as $A$ and the coefficient of $(\frac{1}{\varsigma})^{\beta_{2}}$ as $B$ in equation \eqref{eqn:30}. Now, we will check if there exists a root in $ex \in [0, l_{th}]$ or not.

\begin{itemize}[leftmargin=*]
    \item \textbf{Case 1: }As $ex \to 0$, then $L^{+}(\pi) \to 0$ and $\varsigma \to 0$. As we know $\beta_{1} > 1$ and $\beta_{2} < 0$, thus $(\frac{1}{\varsigma})^{\beta_{2}} \to 0$ and $(\frac{1}{\varsigma})^{\beta_{1}} \to \infty$. Since we are considering the case when $\pi \to 0$, hence the payoff an \texttt{end-user} gets from the \texttt{system} will be approximately equal to zero. Thus, in the first part of $A$, $\lambda(L^{+}(\pi), \pi) \to 0$ and hence $\lambda^{'}(L^{+}(\pi) \to 0$. Therefore, $\frac{\beta_{2} \lambda(L^{+}(\pi), \pi) - \lambda^{'}(L^{+}(\pi), \pi) L^{+}(\pi)} {(\beta_{2} - \beta_{1})} = 0$. For the second term of $A$ i.e. $+\frac{P (1-\beta_{2}) L^{+}(\pi)}{(r-\mu) (\beta_{2} - \beta_{1})}$, the $numerator > 0$ and $denominator < 0$. As a result, in this case, $\Lambda (\varsigma) \to -\infty$.
    \item \textbf{Case 2: } As $ex \to u_{th}$, then $L^{+}(\pi) \to l_{th}$ and $\varsigma \to 1$. If you put these values in equation \eqref{eqn:24}, we will see that $\Lambda (\varsigma) > 0$. 
\end{itemize}

Hence, $\Lambda (\varsigma)$ has at-least one root which lies in range $ex \in [0, u_{th}]$.

\noindent
\textbf{Proof of uniqueness: } Now, we will show that only one root exists for the pair of equations defined above by proving that $\partial \Lambda (\varsigma)$ or $\Lambda_{L^{+}}(\varsigma)$ is increasing in the interval $L^{+}(\pi) \in [0, l_{th}]$ (we could have also taken $L^{-}(\pi)$). Using chain of rule: - 
$\Lambda_{L^{+}}(\varsigma)$ = $\Lambda_{L^{+}}(\varsigma)$ + $\Lambda_{\varsigma}(\varsigma) \times \varsigma^{'}$. Hence, $\Lambda (\varsigma)$ will be increasing in the interval or will have a unique root in $[0, l_{th}]$ if  $\Lambda_{L^{+}}(\varsigma) > 0$ and $\Lambda_{\varsigma}(\varsigma) \times \varsigma^{'} > 0$. \\

\textbf{Claim i: } For all $ex, \pi$, we have $\Lambda_{L^{+}}(\varsigma) > 0$.
We have, 

{\scriptsize
\begin{equation}
\begin{split}
\Lambda (\varsigma) = \left[\frac{\beta_{2} \lambda(L^{+}(\pi), \pi) - \lambda^{'}(L^{+}(\pi), \pi) L^{+}(\pi)} {(\beta_{2} - \beta_{1})} +\frac{P (1-\beta_{2}) L^{+}(\pi)}{(r-\mu) (\beta_{2} - \beta_{1})} \right] \\ \times (\frac{1}{\varsigma})^{\beta_{1}} + \\ \left[\frac{\beta_{1} \lambda(L^{+}(\pi), \pi) - \lambda^{'}(L^{+}(\pi), \pi) L^{+}(\pi)} {(\beta_{1} - \beta_{2})} +\frac{P (1-\beta_{1}) L^{+}(\pi)}{(r-\mu) (\beta_{1} - \beta_{2})} \right] \\ \times (\frac{1}{\varsigma})^{\beta_{2}}
\end{split}
\end{equation}
}

Rearranging the terms will give us:
{\scriptsize
\begin{equation}
\begin{split}
\Lambda (\varsigma) = \frac{1}{\beta_{1} - \beta_{2}} \\ \left[ \beta_{1} \lambda(L^{+}(\pi), \pi) - \lambda^{'}(L^{+}(\pi), \pi) L^{+}(\pi) + \frac{P (1-\beta_{1}) L^{+}(\pi)}{(r-\mu)} \right] \\ \times \frac{1}{\varsigma^{\beta_{2}}} -\frac{1}{\beta_{1} - \beta_{2}} \\ \left[ \beta_{2} \lambda(L^{+}(\pi), \pi) - \lambda^{'}(L^{+}(\pi), \pi) L^{+}(\pi) + \frac{P (1-\beta_{2}) L^{+}(\pi)}{(r-\mu)}\right] \\ \times \frac{1}{\varsigma^{\beta_{1}}} \label{eqn:31}
\end{split}
\end{equation}
}

Now taking derivative of equation \eqref{eqn:31} with respect to $L^{+}(\pi)$.
Taking $L^{+}(\pi) = L^{+}$ and $\lambda(L^{+}(\pi), \pi) = \lambda$ for simplicity.

{\scriptsize
$$ \Lambda_{L^{+}} (\varsigma) = \frac{1}{\beta_{1} - \beta_{2}}\left[ \beta_{1} \lambda^{'} - \lambda^{''}L^{+} - \lambda^{'} + \frac{P (1-\beta_{1}}{r-\mu}\right] \times \frac{1}{\varsigma^{\beta_{2}}} $$
$$\frac{1}{\beta_{1} - \beta_{2}}\left[- \beta_{2} \lambda^{'} + \lambda^{''}L^{+} + \lambda^{'} - \frac{P (1-\beta_{2}}{r-\mu}\right] \times \frac{1}{\varsigma^{\beta_{1}}}$$
}

By rearranging the terms, we get
{\scriptsize
$$ \Lambda_{L^{+}} (\varsigma) = \frac{1}{\beta_{1} - \beta_{2}}\left[ (\beta_{1} -1) ( \lambda^{'} - \frac{P}{r-\mu} ) - \lambda^{''} L^{+} \right] \times \frac{1}{\varsigma^{\beta_{2}}} $$
\begin{equation}
\frac{1}{\beta_{1} - \beta_{2}}\left[ \lambda^{''}L^{+} + (1-\beta_{2}) ( \lambda^{'} - \frac{P}{r-\mu})\right] \times \frac{1}{\varsigma^{\beta_{1}}} \label{eqn:32}
\end{equation}
}
As one can notice, $\lambda^{'} - \frac{sig}{r-\mu}$ is part of equation \eqref{eqn:11} or \eqref{eqn:16}, hence positive. Thus, $\Lambda_{L^{+}} (\varsigma) > 0$. \\

\textbf{Claim ii: } For all $ex, \pi$, we have $\Lambda_{\varsigma}(\varsigma) \times \varsigma^{'} > 0$.

We know that $\varsigma = L^{+}(z)/ L^{-}(z)$, thus $\varsigma_{L^{+}} > 0$.

{\scriptsize
$$\Lambda_{\varsigma}(\varsigma) = \frac{-\beta_{2}}{\beta_{1} - \beta_{2}}\left[ \beta_{1} \lambda - \lambda^{'} L^{+} + \frac{P (1-\beta_{1}) L^{+}}{(r-\mu)} \right] \times \frac{1}{\varsigma^{\beta_{2} + 1}}$$
\begin{equation}
-\frac{\beta_{1}}{\beta_{1} - \beta_{2}} \left[ \beta_{2} \lambda - \lambda^{'} L^{+} + \frac{P (1-\beta_{2}) L^{+}}{(r-\mu)}\right] \times \frac{1}{\varsigma^{\beta_{1}+1}}  \label{eqn:33}
\end{equation}
}

Rearranging equation \eqref{eqn:33} gives us, 
{\scriptsize
$$\Lambda_{\varsigma}(\varsigma) = \frac{-\beta_{2}}{\beta_{1} - \beta_{2}}\left[ \beta_{1} (\lambda - \frac{P L^{+}}{r-\mu}) - L^{+}(\lambda^{'} - \frac{sig}{r-\mu})\right] \times \frac{1}{\varsigma^{\beta_{2} + 1}}$$

\begin{equation}
-\frac{\beta_{1}}{\beta_{1} - \beta_{2}} \left[ L^{+} (\lambda^{'} -\frac{sig}{r-\mu}) + \beta_{2}(\frac{P L^{+}}{r-\mu} - \lambda)\right] \times \frac{1}{\varsigma^{\beta_{1}+1}}  \label{eqn:34}
\end{equation}
}

For $\Lambda_{\varsigma}(\varsigma) > 0$, we must have, 

{\scriptsize
$$ \frac{\beta_{2}}{\varsigma^{\beta_{2} + 1}}\left[L^{+}(\lambda^{'} - \frac{P}{r-\mu}) - \beta_{1} (\lambda - \frac{P L^{+}}{r-\mu})\right]$$
$$ \geq  \frac{\beta_{1}}{\varsigma^{\beta_{1} + 1}}\left[ \beta_{2}(\lambda - \frac{P L^{+}}{r-\mu}) - L^{+} (\lambda^{'} -\frac{P}{r-\mu})\right]$$

}

Hence, if the above condition is true, we will have a unique root of $\Lambda (\varsigma) $which exists in the range $[0, l_{th}]$. Hence proved. 

\end{proof}

%\bibliography{uai2022-template}

%\appendix

\end{document}